\documentclass[letterpaper,twocolumn,10pt]{article}
\usepackage{usenix-2020-09}
\usepackage{tikz}
\usepackage{amsmath}
\usepackage{amssymb}
\usepackage{cleveref}
\usepackage{xcolor}
\usepackage{subcaption}
\usepackage{adjustbox}
\usepackage{pgfplots}
\usepackage{xspace}
\usepackage{algorithm}
\usepackage{algorithmic}
\usepackage[title]{appendix}
\usepackage{tabularx}
\usepackage[subtle]{savetrees}
\usepackage{multirow}
\usepackage{booktabs}
\usepackage{relsize}
\usepackage{cancel}
\usepackage{amsthm}
\usepackage{mathtools}
\usepackage{titlesec}
\usepackage{titling}

\newtheorem{theorem}{Theorem}[section]

\theoremstyle{definition}

\theoremstyle{remark}

\def\compactify{\itemsep=0pt \topsep=0pt \partopsep=0pt \parsep=0pt}
\let\latexusecounter=\usecounter

\newenvironment{CompactEnumerate}
  {\def\usecounter{\compactify\latexusecounter}
   \begin{enumerate}}
  {\end{enumerate}\let\usecounter=\latexusecounter}

\pgfplotsset{compat=newest}
\pgfplotsset{
    colormap={whiteblack}{gray(0cm)=(1); gray(0.5cm)=(0.2); gray(1cm)=(0)}
}

\newcommand{\CausalSim}{CausalSim\xspace}

\newcommand{\bOne}{{\mathbf 1}}
\newcommand{\be}{{\mathbf e}}

\usepackage[acronym]{glossaries}
\glsdisablehyper
\newacronym{ml}{ML}{Machine Learning}
\newacronym[plural=NNs, firstplural=Neural Networks (NNs)]{nn}{NN}{Neural Network}
\newacronym{mdp}{MDP}{Markov Decision Process}
\newacronym{pomdp}{POMDP}{Partially Observable Markov Decision Process}
\newacronym{mp}{MP}{Markov Process}
\newacronym{pomp}{POMP}{Partially Observable Markov Process}
\newacronym{rl}{RL}{Reinforcement Learning}
\newacronym{a2c}{A2C}{Advantage Actor Critic}
\newacronym{gae}{GAE}{Generalized Advantage Estimation}
\newacronym{qoe}{QoE}{Quality of Experience}
\newacronym{dnn}{DNN}{Deep Neural Network}
\newacronym{abr}{ABR}{Adaptive Bit Rate}
\newacronym{bba}{BBA}{Buffer-Based Algorithm}
\newacronym{rct}{RCT}{Randomized Control Trial}
\newacronym{scm}{SCM}{Structural Causal Model}
\newacronym{rtt}{RTT}{Round Trip Time}
\newacronym{tcp}{TCP}{Transmission Control Protocol}
\newacronym{mlp}{MLP}{Multi Layer Perceptron}
\newacronym{relu}{ReLU}{Rectified Linear Unit}
\newacronym{emd}{EMD}{Earth Mover Distance}
\newacronym{cdf}{CDF}{Cumulative Distribution Function}
\newacronym{ood}{OOD}{Out of Distribution}
\newacronym{pcc}{PCC}{Pearson Correlation Coefficient}
\newacronym{aimd}{AIMD}{Additive Increase - Multiplicative Decrease}
\newacronym{mse}{MSE}{Mean Squared Error}
\newacronym{wise}{WISE}{What-If Scenario Evaluator}
\newacronym{cfa}{CFA}{Critical Feature Analytics}
\newacronym{mape}{MAPE}{Mean Absolute Percentage Error}
\newacronym{adda}{ADDA}{Adversarial Discriminative Domain Adaptation}
\newacronym{dag}{DAG}{Directed Acyclic Graph}
\newacronym{cdn}{CDN}{Content Delivery Network}
\newacronym{hmm}{HMM}{Hidden Markov Model}

\usepackage{listings}
\usepackage{times}
\begin{document}

\setlength{\droptitle}{-1.2cm}

\date{\vspace{-8mm}}

\title{\Large \bf \CausalSim: A Causal Framework for Unbiased Trace-Driven Simulation}

\author{
{\rm Abdullah Alomar$^*$} \\
MIT\\
\texttt{aalomar@mit.edu}
\and
{\rm Pouya Hamadanian$^*$} \\
MIT\\
\texttt{pouyah@mit.edu}
\and
{\rm Arash Nasr-Esfahany$^*$}\\
MIT\\
\texttt{arashne@mit.edu}
\and
{\rm Anish Agarwal}\\
MIT\\
\texttt{anish90@mit.edu}
\and
{\rm Mohammad Alizadeh}\\
MIT \\
\texttt{alizadeh@mit.edu}
\and
{\rm Devavrat Shah}\\
MIT\\
\texttt{devavrat@mit.edu}
}

\maketitle

\begin{abstract}
\def\thefootnote{*}\footnotetext{Equal contribution}\def\thefootnote{\arabic{footnote}}
We present \CausalSim, a causal framework for unbiased trace-driven simulation. Current trace-driven simulators assume that the interventions being simulated (e.g., a new algorithm) would not affect the validity of the traces. However, real-world traces are often biased by the choices algorithms make during trace collection, and hence replaying traces under an intervention may lead to incorrect results. \CausalSim addresses this challenge by learning a causal model of the system dynamics and latent factors capturing the underlying system conditions during trace collection. It learns these models using an initial randomized control trial (RCT) under a fixed set of algorithms, and then applies them to remove biases from trace data when simulating new algorithms.

Key to \CausalSim is mapping unbiased trace-driven simulation to a tensor completion problem with extremely sparse observations. By exploiting a basic distributional invariance property present in RCT data, \CausalSim enables a novel tensor completion method despite the sparsity of observations. 
Our extensive evaluation of \CausalSim on both real and synthetic datasets, including more than ten months of real data from the Puffer video streaming system shows it improves simulation accuracy, reducing errors by 53\% and 61\% on average compared to expert-designed and supervised learning baselines. Moreover, \CausalSim provides markedly different insights about ABR algorithms compared to the biased baseline simulator, which we validate with a real deployment.
\end{abstract}

\section{Introduction}
\label{sec:intro}
{\em Causa Latet Vis Est Notissima \,--\, The cause is hidden, but the result is known. (Ovid: Metamorphoses IV, 287)}

\smallskip
Trace-driven simulation is a widely used method for evaluating new ideas in systems. 
In contrast to full-system simulation (e.g., NS3~\cite{ns3}), which requires detailed knowledge of system characteristics (e.g., topology, traffic patterns, hardware details, etc.), 
trace-driven simulation does not model all components of a system.
Instead, it focuses on simulating one (or a few) components of interest, where we wish to experiment with an {\em intervention}, e.g., a new design, algorithm, or architectural choice. To account for the effect of the remaining components that are not simulated, we collect a trace capturing their behavior and replay it while simulating the component of interest with the proposed intervention.

The key assumption here is that the interventions would not affect the trace being replayed, which we refer to as the \emph{exogenous trace} assumption.
If this assumption does not hold, replaying the trace is invalid and could lead to incorrect simulation results. This problem has been referred to as {\em bias} in trace-driven (or data-driven) simulation~\cite{bartulovic2017biases, jiang2017unleashing}.

It is difficult to guarantee the exogenous trace assumption in traces collected from real-world systems. 
Consider, for example, trace-driven simulation of adaptive bitrate (ABR) algorithms~\cite{bba, mpc, bola, pensieve}.
It is common to use network throughput traces from real video streaming sessions on Internet paths~\cite{FESTIVE, mpc}.
However, the throughput achieved when the player downloads a video chunk is caused by certain {\em latent} properties of the network path (e.g., the underlying bottleneck capacity, the number and type of competing flows, etc.), as well as the particular choices made by the ABR algorithm (the bitrate chosen for each chunk). 
In other words, the trace data reflects the combined effect of these two causes and is biased by the ABR algorithms used during trace collection.
To simulate a new algorithm, we need to tease apart the effect of the two causes, and predict how the trace would have changed under the decisions of the new algorithm.

We present \CausalSim, a causal framework for unbiased trace-driven simulation.
\CausalSim relaxes the \emph{exogenous trace} assumption by
explicitly modeling the fact that interventions can affect trace data.
Using traces collected from a {\em randomized control trial} (RCT) under a fixed set of algorithms, it infers both the latent factors capturing the underlying conditions of the system and a causal model of its dynamics, including the unknown relationship between latents, algorithm decisions, and observed trace data. To simulate a new algorithm, \CausalSim first estimates the latent factors at every time step of each trace. Then, it uses the estimated latent factors to predict the alternate evolution of the trace, actions, and observed variables of the component of interest, under the same latent conditions that were present when the trace was collected.
This two-step process allows \CausalSim to remove the bias in the trace data when simulating new algorithms.

\CausalSim provides two benefits: (\emph{i}) it improves the accuracy of trace-driven simulation when the intervention could affect (in possibly subtle ways) the trace data; 
(\emph{ii}) it enables trace-driven simulation of systems where defining an exogenous trace is not possible and therefore standard trace-driven simulation is not applicable. We evaluate both settings in this paper, by simulating ABR and heterogeneous server load balancing algorithms as examples for cases (\emph{i}) and (\emph{ii}) respectively. 

\CausalSim requires training data from an RCT. Large network operators have increasingly invested in RCT infrastructure to evaluate new ideas, but due to their low  throughput and risk of disruptions or SLA violations~\cite{krishnan2013video}, they can afford to evaluate only a fraction of proposed ideas in RCTs. \CausalSim greatly extends the utility of RCT data by learning a model that can simulate a wide range of algorithms using traces from a fixed set of algorithms. Periodically or whenever an operator believes the underlying system characteristics  have changed significantly, they can collect fresh data using an RCT (again, with the same fixed set of algorithms) to retrain \CausalSim.

\CausalSim's design begins with the observation that unbiased trace-driven simulation can be viewed as a matrix (or tensor) completion problem~\cite{murphy-completion, agarwal2021synthetic}.
Consider a matrix $M$ of traces (it is a tensor if traces are higher dimensional), with rows corresponding to possible actions and columns corresponding to different time steps in the trace data.
For each column, the entry for one action is ``revealed''; all other entries are missing.
Our task can be viewed as recovering the missing entries.

A significant body of work has shown that it is possible to recover a matrix from sparse observations under certain assumptions about the matrix and the pattern of missing data. Roughly speaking, the typical assumptions that make recovery feasible are that the matrix has low rank, the entries revealed are chosen at random, and that enough entries are revealed.
Low-rank structure is prevalent in many real-world problems~\cite{udell2019big} and has also been observed in network measurement data~\cite{traffic1, traffic2, traffic3, traffic4}.
But unfortunately the other two assumptions do not hold in our problem.
As we detail in \S\ref{subsec:MC_discussion}, one observed entry per column is below the information-theoretic bound for low-rank matrix completion (even for rank $r=1$).
Moreover, not only are the entries revealed in our problem not random, they depend on other entries of the matrix, since the actions are being taken by algorithms based on observed variables.

To overcome these challenges, \CausalSim exploits two key insights.
First, it assumes a causal model (\S\ref{sec:causality}) where the latent factors are exogenous and are not affected by the interventions we want to simulate in the component of interest. This {\em exogenous latent} assumption relaxes (and is therefore implied by) the exogenous trace assumption in standard trace-driven simulation. 
For example, in ABR, it says that underlying factors like the bottleneck link speed on a network path are not affected by a user's ABR algorithm, whereas ABR decisions can impact the trace that user {\em observes} (i.e., the achieved throughput).

Second, \CausalSim uses a basic property of trace data collected via an RCT.
Since the assignment of an algorithm to a trace is completely random in an RCT, the distribution of latent factors should be the same for the traces obtained using different algorithms, i.e., the latent distribution is {\em invariant} to the algorithm. 
We provide conditions on the RCT data (e.g., in terms of the number and diversity of algorithms) that guarantee recoverability of the low-rank matrix  using this invariance property (\S\ref{sub:MC+invar}), and we operationalize this idea in a practical learning method that exploits the invariance using an adversarial neural network training technique (\S\ref{sec:solution}). 

We evaluate \CausalSim on two use cases, ABR and server load balancing, with both real-world and synthetic datasets, and further verify \CausalSim's predictions with a test in the wild on the Puffer~\cite{puffer} video streaming testbed.
Our main findings are:
\begin{CompactEnumerate}
\item We use \CausalSim to debug and improve an ABR algorithm, BOLA1~\cite{bola, marx2020implementing}. In a ten month experiment on Puffer~\cite{puffer}, BOLA1 exhibited high stalling compared to  BBA~\cite{bba}, with slightly better quality.
Using \CausalSim, we tune BOLA1's parameters via Bayesian Optimization and deploy our improved version on Puffer. We show that it improves the stall rate of this well-known algorithm by $2.6\times$, achieving $0.7\times$ the stall rate of BBA with similar perceptual quality. The expert-designed baseline simulator that ignores bias predicts the exact opposite: that the new variant should stall $1.34\times$ the stall rate of BBA. This case study shows that removing bias is crucial to draw accurate conclusions from trace-driven simulation. 

\item Evaluation of \CausalSim on more than ten months of real data from  Puffer shows that \CausalSim's error in stall rate prediction is bounded to 28\%, while expert-designed and standard supervised learning baselines have errors in the range of 49--68\% and 29--187\% respectively. Similar observations are also made for perceptual quality metrics and buffer occupancy levels.
\item \CausalSim opens up new avenues to apply trace-driven simulation to systems where the exogenous trace assumption is invalid. Using a synthetic environment modeling a heterogeneous server load balancing problem, we show how \CausalSim reduces average simulation error by $5.1\times$, a stark improvement compared to a baseline simulator with a median error of 124.3\%.
\end{CompactEnumerate}

This work does not raise any ethical issues. Our code is available at \href{https://github.com/CausalSim/Unbiased-Trace-Driven-Simulation}{https://github.com/CausalSim/Unbiased-Trace-Driven-Simulation}.
\section{Motivation}
\label{sec:motivation}

\subsection{Bias in Trace-Driven Simulation}
Trace-driven simulation is a widely used technique to design and evaluate systems. Unlike full-system simulation, it focuses on simulating one (or a few) components of the system while capturing the effect of remaining components by replaying a trace. For example, to simulate new ABR algorithms, it is common to replay network throughput traces from real Internet paths in a simulator modeling only the video player/server. 

\begin{figure}
    \hspace{0.05\linewidth}
    \begin{subfigure}{0.4\linewidth}
        \centering
        \includegraphics[width=0.89\linewidth]{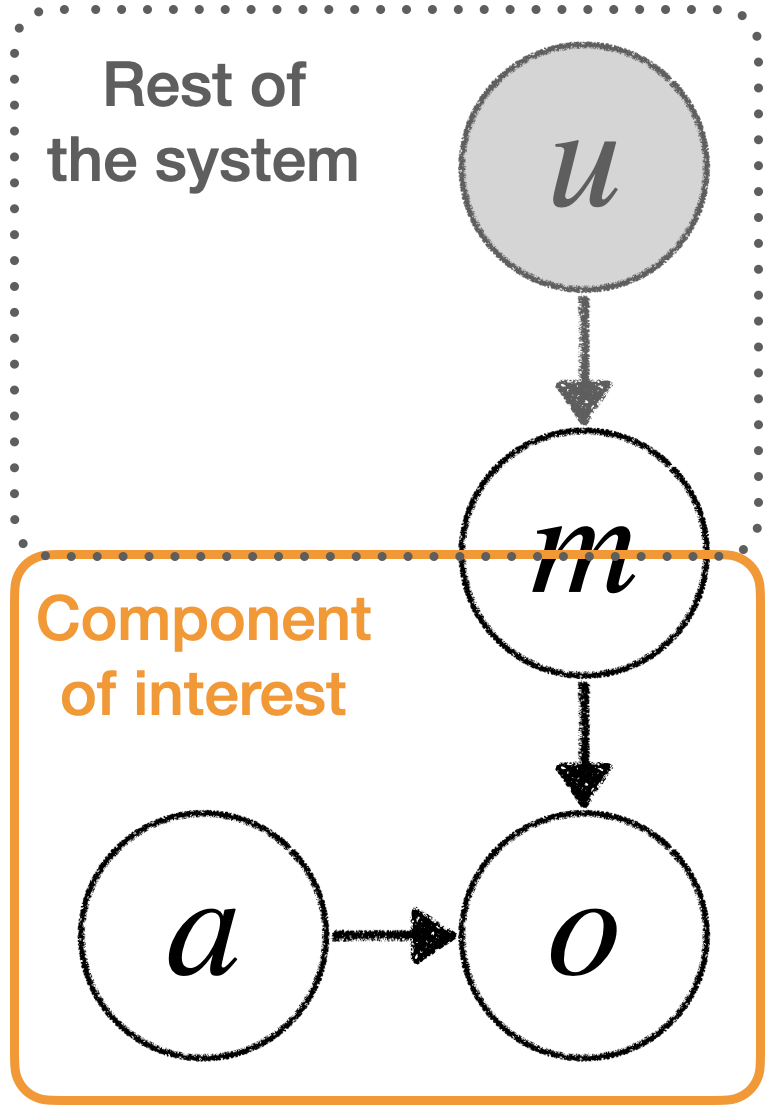}
        \caption{Trace-driven simulation}
        \label{fig:trace-driven-dag}
    \end{subfigure}
    \hspace{0.1\linewidth}
    \begin{subfigure}{0.4\linewidth}
        \centering
        \includegraphics[width=0.89\linewidth]{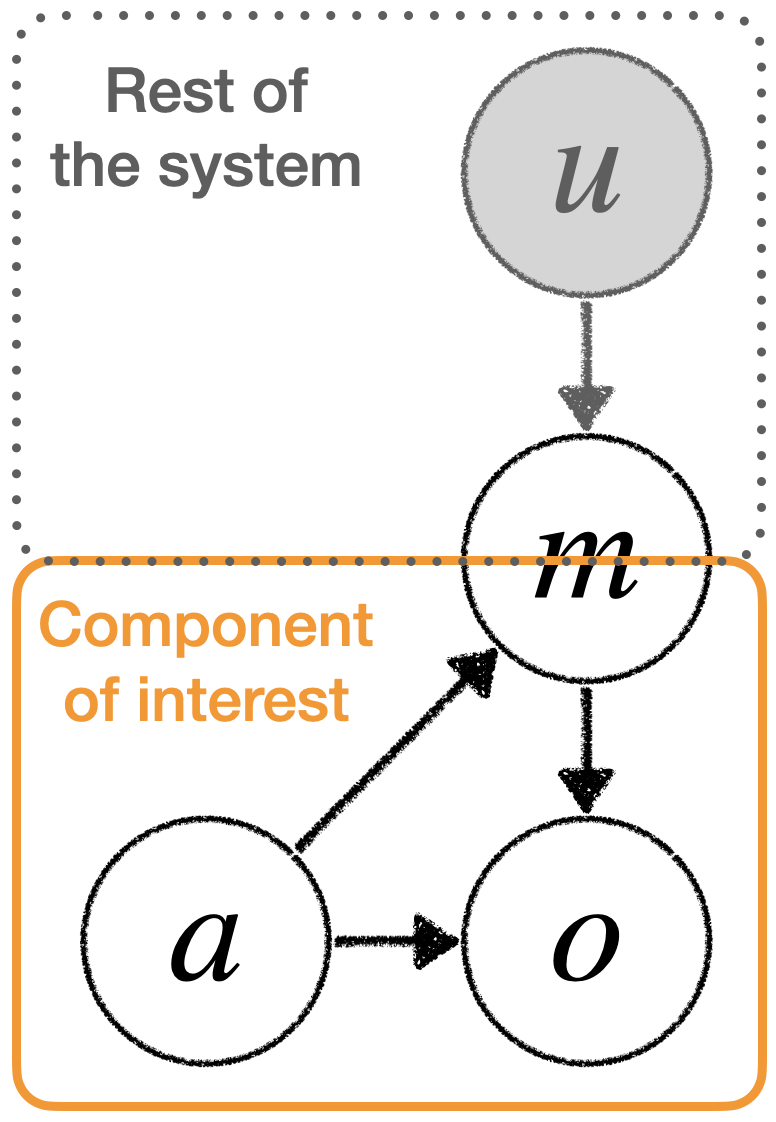}
        \caption{\CausalSim}
        \label{fig:causalsim-dag}
    \end{subfigure}
    \caption[\CausalSim relaxes the \emph{exogenous trace} assumption in standard trace-driven simulation.]{\CausalSim relaxes the \emph{exogenous trace} assumption in standard trace-driven simulation.\protect\footnotemark}
    \label{fig:dags}
\end{figure}

\footnotetext{In general, $a$ and $u$ can be correlated. For example, they  can both depend on prior latent conditions of the system. In ABR, for instance, recent latent path conditions are correlated with current path conditions ($u$), and also affect the action taken by the ABR algorithm ($a$). Correlation of $a$ and $u$, however, does not imply a causal relationship between them. In particular, our model assumes {\em exogenous latents}, i.e. $a$ does not affect $u$.}

As we alluded to earlier, the key assumption here is that the interventions being simulated would not affect the trace being replayed; otherwise, replaying the trace would be invalid. We refer to this as the \emph{exogenous trace} assumption, and it is central to standard trace-driven simulation. \Cref{fig:trace-driven-dag} is a visual depiction of the exogenous trace assumption.
In the figure, $a$ represents the intervention we want to simulate; for example, the actions taken by a new algorithm. 
$o$ is the observed state of the component being simulated. $u$ represents the {\em latent} state of the rest of the system, which we do not observe or simulate. Finally, $m$ is the trace, which captures the behavior of the other components.\footnote{Variables in Fig.~\ref{fig:trace-driven-dag} can be multidimensional and vary with time.} 
The existence of each edge represents a causal effect. For example, the trace $m$ and intervention $a$ both affect $o$. Note the absence of the edge from $a$ to $m$, which implies that the intervention cannot affect the trace (the exogenous trace assumption).

The simulator designer must define the trace carefully to meet this assumption.
But what happens if it does not hold, i.e., there exists an edge from $a$ to $m$ (as in \Cref{fig:causalsim-dag})?
Ignoring the violation of \emph{exogenous trace} assumption leads to biased simulation outcomes, as we will see next. 

\subsection{An Example Using Real-world Traces} 

In this section, we use more than ten months of real-world data from Puffer~\cite{puffer}, a recently deployed system for experimenting with video streaming protocols, to illustrate the issue of bias in trace-driven simulation. 

Puffer collects data from a continual \gls{rct} that tests several \gls{abr} algorithms. 
In the period of interest (July 27, 2020 -- June 2, 2021), the tested algorithms include \gls{bba}~\cite{bba}, two versions of BOLA-BASIC (henceforth called BOLA)~\cite{bola}\footnote{BOLA1 and BOLA2 are variations on BOLA adjusted to target the SSIM quality metric instead of bitrate~\cite{marx2020implementing}. They pursue different objective functions and use different principles for hyperparameter adjustment.}, and two versions of an algorithm called Fugu developed by the Puffer authors. 
The dataset includes more than $56$ million chunk downloads from more than $230$ thousand streaming sessions, 
totaling $3.5$ years of streamed videos. For each streaming session, it provides logs of the chosen chunk sizes, available chunk sizes, achieved chunk download throughputs, and playback buffer levels.\footnote{We use `slow stream' logs (by Puffer's definition, streams with TCP delivery rates below 6Mbps) available on the Puffer website~\cite{puffer_daily}.}  

Consider a typical trace-driven simulation scenario, where we wish to simulate a new ABR algorithm using traces from previous video streaming sessions. We define such a task on the Puffer data as follows. We let one of the algorithms, say BBA, be the algorithm that we wish to simulate. We leave out the data for this algorithm and ask whether it is possible to predict its performance using the other algorithms' traces. %
In evaluating a new ABR algorithm, we may be interested in various performance measurements, e.g.\ buffer occupancy, rebuffering rate, chosen bitrates, etc. Here, we focus on predicting the behavior of playback buffer occupancy, which is one of the key indicators of an ABR algorithm's behavior~\cite{bba}.

The goal of trace-driven simulation is to predict the trajectory of the system (e.g., buffer, bitrates, etc.) for one algorithm in {\em the same underlying conditions} that were present when a trace was collected using a different algorithm. When simulating algorithm $B$ based on a trace collected using algorithm $A$, we will refer to $A$ as the ``source'' algorithm and to $B$ as the ``target'' algorithm.

It is generally not possible to evaluate the accuracy of individual simulated trajectories using real-world data, because we do not have ground truth trajectories for the target algorithm under the same exact network conditions that were present when running the source algorithm. However, since the Puffer data was obtained using an RCT, we can evaluate predictions about {\em distributional} properties of the target algorithm, such as the distribution of the buffer occupancy achieved by the algorithm over the population of network paths present in the RCT. 

To summarize, our task is: 
\emph{predict the distribution of the buffer occupancy for the users assigned to BBA (the target algorithm) in the Puffer dataset, using only the data from the other (source) algorithms.} 

\begin{figure}
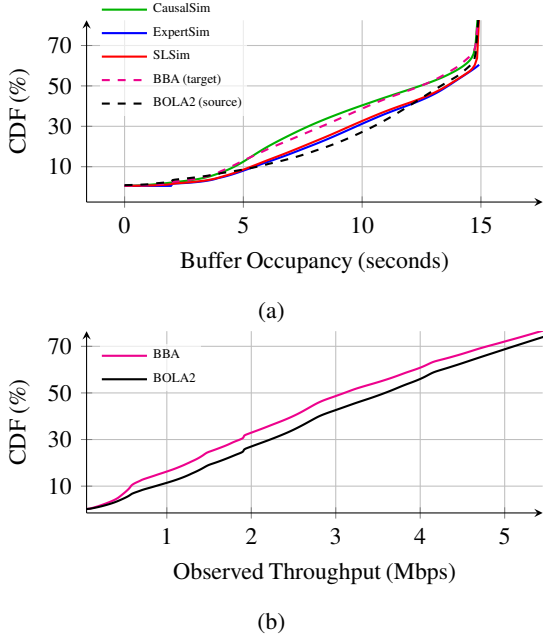

\begin{subfigure}{0.43\textwidth}
    \centering
    \begin{tikzpicture}
        \begin{axis}[height=4cm,
                     width=\linewidth,
                     axis lines=left,
                     xlabel={Buffer Occupancy (seconds)},
                     ylabel={CDF (\%)},
                     font=\small,
                     legend columns=1,
                     enlargelimits=true,
                     legend style={at={(0.03,0.8)}, 
                          anchor=west, fill=white, fill opacity=0.7, draw opacity=1, 
                          text opacity=1, draw=none, font=\tiny, inner sep=0},
                          legend cell align={left},
                     xmax=16,
                     xmin=0,
                     ymax=75, 
                     ymin=-0,
                     ytick={10, 30, 50, 70, 90},
                     ylabel near ticks,
                     xlabel near ticks,
                     grid=both
                     ]
            \input{figures/raw_data/appendix/new_puffer_left_out/new_linear_bba_bola_basic_v2}
            \legend{
            \CausalSim, 
            ExpertSim, 
            SLSim, 
            BBA (target), 
            BOLA2 (source)
            };
        \end{axis}
    \end{tikzpicture}
    \caption{}
    \label{fig:motiv:buffer}
  \end{subfigure}
  \begin{subfigure}{0.43\textwidth}
    \centering
    \begin{tikzpicture}
        \begin{axis}[height=4cm,
                     width=\linewidth,
                     axis lines=left,
                     xlabel={Observed Throughput (Mbps)},
                     ylabel={CDF (\%)},
                     font=\small,
                     legend columns=1,
                     enlargelimits=true,
                     legend style={at={(0.03,0.8)}, 
                          anchor=west, fill=white, fill opacity=0.7, draw opacity=1, 
                          text opacity=1, draw=none, font=\tiny, inner sep=0},
                          legend cell align={left},
                     xmax=5,
                     xmin=0.5,
                     ymax=70, 
                     ymin=5,
                     ytick={10, 30, 50, 70, 90},
                     ylabel near ticks,
                     xlabel near ticks,
                     grid=both
                     ]
            \input{figures/raw_data/chat_motivation_v2}
            \legend{
            BBA 
            ,BOLA2
            };
        \end{axis}
    \end{tikzpicture}
    \caption{}
    \label{fig:motiv:chat}
  \end{subfigure}
     \vspace{-0.03\linewidth}
  \caption{\textbf{(a)} \CausalSim is accurate in predicting buffer level distribution of BBA users, while baseline simulators' predictions are similar to BOLA2 users.
  \textbf{(b)} Distribution of achieved throughput is different in BBA and BOLA2 users.
}
\end{figure}

\subsubsection{Simulation via Expert Modeling (ExpertSim)}
\label{subsub:expert_sim}

As our first strawman, we build a simple trace-driven simulator (ExpertSim) using our knowledge of how an \gls{abr} system works.
ExpertSim models the playback buffer dynamics for each {\em step}, where a step corresponds to one ABR decision and the download of a single video chunk. 
Let $\hat{c}_t$ be the throughput achieved in step $t$ (for the $t^{th}$ chunk) of a particular video streaming session using, say, the BOLA2 algorithm. 
To simulate BBA for the same user, ExpertSim assumes that the user would achieve the same throughput $\hat{c}_t$ in each step under the BBA algorithm as well. In other words, it assumes that ABR decisions do not affect the observed network throughput (the exogenous trace assumption). Under this assumption, ExpertSim models the evolution of the video playback buffer as follows. Let $b_t$ be the buffer level at the beginning of step $t$ (before the download of chunk $t$), $r_t$ be the bitrate chosen in step $t$, and $s_t$ be the size of the $t^{\text{th}}$ chunk implied by the chosen bitrate. Then the buffer at the end of step $t$ is derived as: $b_{t+1} = \max(0, b_t - s_t/\hat{c}_t) + T$, where $T$ is the chunk 
duration.\footnote{The complete buffer dynamic equation is slightly more complex to handle cases with full buffers. Refer to \S\ref{app:abr_synth:dynamics} in the appendix for further clarification.}
Although simple, the assumption that throughput is an exogenous property of a network path is common in modelling ABR protocols. For example, both FastMPC~\cite{mpc} and FESTIVE~\cite{FESTIVE} assume that the observed throughput does not depend on the chosen bitrate.

\Cref{fig:motiv:buffer} shows the true distribution of buffer level for BOLA2 and BBA users in the Puffer dataset (the two dashed lines), as well as the distribution {\em predicted} by running BBA on the traces collected from BOLA2 users using ExpertSim (solid blue line).
The predictions are inaccurate: the buffer distribution generated by ExpertSim is more similar to the buffer distribution of BOLA2 users (the source algorithm) than the buffer distribution of BBA users (the target algorithm).

\subsubsection{Simulation via Supervised Learning (SLSim)}
\label{subsub:data-driven}
Perhaps the simple model of buffer dynamics in ExpertSim does not accurately reflect the actual system behavior. 
As a next attempt, we turn to machine learning and try to learn the system dynamics from data.
Specifically, we use supervised learning to train a \gls{nn} that models the step-wise dynamics of the system. This fully connected \gls{nn} includes 2 hidden layers, each with 128 ReLU activated neurons.
For each timestep $t$, the \gls{nn} takes as input the buffer level before downloading the $t^{\text{th}}$ chunk $b_t$, the achieved throughput $\hat{c}_t$ for chunk $t$, and the chunk size $s_t$ (which depends on the birate chosen by ABR). The \gls{nn} outputs the download time of the $t^{th}$ chunk, and the resulting buffer level $b_{t+1}$. We train the \gls{nn} to minimize the prediction error on our dataset. To avoid information leaking, we exclude the logs for BBA from the training data.

\Cref{fig:motiv:buffer} shows the predicted buffer level distribution via this approach (SLSim) for BBA. 
As with ExpertSim, we use the traces collected from BOLA2 users as the source algorithm.
The results are similar to ExpertSim; once again, the predicted buffer distribution is closer to that of BOLA2 than BBA. 

\subsubsection{What Went Wrong?}\label{ssec:minRTT}
To understand the limitations of ExpertSim and SLSim, we plot the distribution of achieved per-chunk throughput for users assigned to BOLA2 and BBA in \Cref{fig:motiv:chat}.
Since algorithm selection is completely random, we would expect inherent network path properties such as bottleneck link capacity to have the same distribution for users assigned to different ABR algorithms.
However, such an invariance should not be expected for achieved throughput, because even on the same path different ABR algorithms could achieve different throughput. 
For example, since congestion control protocols take time to discover available bandwidth (e.g., in slow start) or converge to their fair share rate when competing against other flows, an ABR algorithm that tends to choose lower bitrates (and hence download less data per chunk) may achieve less throughput than an ABR algorithm that picks higher bitrates~\cite{huang2012confused,ABRpitfals}.  
We can see this behavior in the Puffer dataset. The achieved throughput for BOLA2 and BBA is clearly different in \Cref{fig:motiv:chat}.

This confirms that ABR algorithms cause a bias in the measured throughput traces, and the exogenous trace property does not hold. To perform accurate trace-driven simulation, we need to account for this bias when simulating new ABR algorithms.

\subsection{Causal Inference to the Rescue!}

If the traces were the {\em underlying network capacity} when each chunk was downloaded (rather than the achieved throughput), the exogenous trace assumption would hold and our problem would be simple.
First, we would learn the relationship between network capacity and achieved throughput for different ABR actions using our data.  
Then, to simulate BBA for a given trace, we would start with the network capacity at each step of the trace and predict the achieved throughput taking into account the bitrate chosen by BBA in that step. This would then allow us to predict how the buffer evolves.  
This works because unlike achieved throughput, underlying capacity is an exogenous property of a network path and is not affected by the ABR actions.

However, underlying network capacity is a {\em latent} quantity \,---\, we do not observe it in our traces. 
The key challenge is therefore to {\em infer} such latent quantities from observational data. Concretely, in our running example, we wish to estimate the latent factors like network capacity in each step of a trace, using observations such as the bitrate, the chunk size, the achieved throughput, etc.\footnote{For simplicity, we only mention network capacity here, but other latent path conditions like the number of competing flows could also affect achieved throughput and the same reasoning applies to them.
}

Inferring such {\em latent confounders} and using them for counterfactual prediction is the core issue in the field of causal inference~\cite{causality-pearl,causality-peters}. In this paper, we develop \CausalSim, a causal framework for unbiased trace-driven simulation. \CausalSim relaxes the exogenous trace assumption in trace-driven simulation. It explicitly models the fact that interventions can affect trace data (the edge from $a$ to $m$ in \Cref{fig:causalsim-dag}), and infers both the latent factors and a causal model of the system dynamics. This allows \CausalSim to correct for the bias in trace data when simulating an intervention.  
As an illustration, \Cref{fig:motiv:buffer} shows the predicted buffer occupancy distribution when simulating BBA 
on the traces of users assigned to BOLA2, using \CausalSim. \CausalSim matches the ground-truth distribution for BBA much more accurately than the alternatives.
\section{Model and Problem Statement}
\label{sec:causality}

\subsection{Causal Model}
\label{sec:causal-model}

Consider the following discrete-time dynamical model\footnote{This model is similar to a special type of Partially Observable Markovian Decision Processes (POMDPs) in which the unobserved part of the state is exogenous~\cite{mao2018variance}.} corresponding to \Cref{fig:causalsim-dag}:
\begin{align}
\label{eqn:emission}
        m_{t} &= \mathcal{F}_{\text{trace}}(a_t, ~u_t), \\
\label{eqn:system}
     o_{t+1} &= \mathcal{F}_{\text{system}}(o_t, ~m_t, ~a_t).
\end{align}
Here, $t$ denotes the time index, $m_t$ is the trace, $a_t$ is the intervention, $u_t$ is the latent factor, and $o_t$ is the observed state of the component of interest. The function $\mathcal{F}_{\text{trace}}$ models the effect of interventions on the trace (which traditional methods ignore), and $\mathcal{F}_{\text{system}}$ models the dynamics of the component of interest. 
When the intervention changes an algorithm in the component of interest, $a_t$ can be viewed as the action taken by that algorithm at time $t$.

We assume that interventions do not affect the internal state of the rest of the system, i.e., that the latent factors are exogenous. This assumption is implicit in the dynamical system equations, and also visualized in \Cref{fig:causalsim-dag} by the absence of the edge from $a$ to $u$. Note that this is a strict relaxation of the  exogenous trace assumption in standard trace-driven simulation. There, the trace itself is assumed to be unaffected by intervention, which also implies exogenous latent factors. 

In our running \gls{abr} example, we want to simulate the video player and server (components of interest) without precisely modeling the entire network path (the rest of the system). Each time step $t$ corresponds to the download of a new chunk, and $u_t$ represents latent network conditions during that transmission, e.g., bottleneck link speed, number of flows sharing the same network path, type of congestion control used by competing flows, etc.
At each time step, the \gls{abr} algorithm chooses a bitrate $a_t$, which together with $u_t$ generate $m_t$, the \emph{achieved} throughput when downloading a chunk.
Typically, latent network conditions are exogenous factors, beyond the impact of a particular user's actions. 
For instance, the bottleneck link speed and type of congestion control that competing flows use, are not affected by the actions of the \gls{abr} algorithm. 

Note that the achieved throughput depends on the \gls{abr} action as well as the latent network conditions.
\Cref{eqn:emission} captures this relationship and is the source of the bias induced by the ABR algorithm, which we demonstrated in \S\ref{ssec:minRTT}. 

\noindent{\bf When is the model applicable?}
The causal model applies in any trace-driven simulation setting where the trace may be impacted by interventions. Examples include:
\begin{itemize}
    \item Job scheduling, where we wish to simulate a workload's performance under different types of machines. The trace is the job performance (e.g., runtime), interventions are the scheduling decisions, and latent factors are intrinsic properties of each job (e.g., compute intensity) or latent aspects of the machines such as collocated interfering workloads. 
    \item Network simulation, where we wish to simulate how some aspect of network's design (e.g., congestion control, packet scheduling, traffic engineering, etc.) impacts application performance. The trace is an application's traffic pattern, the intervention is the network design, and latent factors are the internals of the application that dictate its traffic demand. 
\end{itemize}
In some cases, like our running ABR example, the exogenous trace assumption may not hold exactly but still be roughly valid.\footnote{Even in these cases, these subtly biased simulations can produce entirely incorrect conclusions (\S\ref{subsec:causalsim-casestudy}).} Here, \CausalSim removes bias and improves simulation accuracy. But in certain problems, ignoring the effect of interventions is meaningless. For example, consider scheduling or load balancing on heterogeneous machines (e.g., with different hardware capabilities). Given a trace of job performance on specific machines, it isn't possible to merely replay the trace for new machine assignments. In such problems, \CausalSim enables trace-driven simulation by explicitly modeling the effect of interventions on the trace.   

\noindent{\bf When is the model invalid?}
Our causal model relaxes the exogenous trace assumption but still requires {\em exogenous latents}, i.e. that the latents are unaffected by the intervention. This won't hold in all systems. For example, we cannot model the effect of network routing policies (e.g., BGP) on observed video streaming throughput in this way, since changing the path would change the latent network conditions that impact a video stream. Another example is simulating the effect of a CPU feature like the branch predictor on instruction throughput. Here, we can't model the state of the instruction/data caches as an exogenous latent factor, since changing the branch predictor can change their internal state significantly. 

Overall, a simulation designer needs to reason about the causal structure of observed and latent quantities to define the appropriate model in the form of Equations~\eqref{eqn:emission} and ~\eqref{eqn:system}. However, the designer does not need to precisely specify the meaning of the latents or the dynamics (the functions $\mathcal{F}_{\text{trace}}$ and $\mathcal{F}_{\text{system}}$). \CausalSim learns both from observational data.

\subsection{Problem Formulation}
\label{sec:sim-is-ce}

We are given $N$ trajectories, collected using $K$ specific policies.\footnote{We use policy and algorithm interchangeably in this paper.}
Let $H_i$ be the length of trajectory $i \in \{1,\dots, N\}$. 
For trajectory $i$, we observe $(m^i_t, o^i_t, a^i_t)_{t=1}^{H_i}$. 
We assume that trajectories are generated using an \gls{rct}, i.e., that each trajectory is assigned to one of the $K$ policies at random.

Our goal is to estimate the observations under an arbitrary given intervention (e.g., a new algorithm) for each of the $N$ trajectories.
Let $\{u^i_t\}_{t=1}^{H_i}$ be the exogenous latent factors for trajectory $i$. 
Formally, for any given trajectory $i$ and given a sequence of actions $\{\tilde{a}^i_t\}_{t=1}^{H_i}$, starting with observation $o^i_1$ and under the same sequence of latent factors $\{u^i_t\}_{t=1}^{H_i}$, we wish to estimate the counterfactual observations $\{\tilde{o}^i_t\}_{t=1}^{H_i}$ that are consistent with \Cref{eqn:emission,eqn:system}. 

This is a counterfactual estimation problem since it requires (i)~estimating {\em latent}
$\{u^i_t\}_{t=1}^{H_i}$ factors for observed trajectory $i$ and using them along with the 
counterfactual actions $\{\tilde{a}^i_t\}_{t=1}^{H_i}$ to predict the counterfactual trace $\{\tilde{m}^i_t\}_{t=1}^{H_i}$ consistent with \Cref{eqn:emission}, and  then
(ii)~using the counterfactual trace and actions to predict counterfactual observations $\{\tilde{o}^i_t\}_{t=1}^{H_i}$ consistent with \Cref{eqn:system}.

For (ii), learning  $\mathcal{F}_{\text{system}}$ is a supervised learning task because its inputs, $(o_t^i, m_t^i, a_t^i)$, and output, $o_{t+1}^i$, are fully observed. 
If $\{u^i_t\}_{t=1}^{H_i}$ was observed, then (i) would also boil down to learning $\mathcal{F}_{\text{trace}}$ 
in a supervised manner.  
{\em It is the lack of observability of $\{u^i_t\}_{t=1}^{H_i}$ that makes our simulation task extremely challenging.
In short, we are left with (i), the task of estimating $\{\tilde{m}^i_t\}_{t=1}^{H_i}$ and learning
$\mathcal{F}_{\text{trace}}$. } 

\section{\CausalSim: Theoretical Insights}
\label{sec:MatrixCompletion}

This section describes the theory behind \CausalSim. We discuss how to operationalize this theory in a practical learning algorithm in \S\ref{sec:solution}.  
We begin by casting counterfactual estimation as a challenging variant of the matrix completion problem~\cite{murphy-completion}.
We then formalize conditions that allow us to complete the matrix using a certain distributional invariance property that is present in data collected in an \gls{rct}.

\subsection{Counterfactual Estimation\\ as Matrix Completion}
\label{sub:matcomp_cf}

Recall from \S\ref{sec:sim-is-ce} the task of estimating the counterfactual trace $\{\tilde{m}^i_t\}_{t=1}^{H_i}$ consistent with \Cref{eqn:emission}. In this section, we pose this task as a variant of the classical matrix completion problem. For simplicity, let action $a_t^i$ be one of the finitely many options $\{1,\dots, A\}$ for some $A \geq 2$.
Imagine an $A$ by $U$ matrix $M$, where rows correspond to 
$A$ potential actions, and columns corresponds to $U = \sum_{i=1}^N H_i$ latent factors ($u^i_t$ for different 
choices of $i$ and $t$) in the dataset.  To order the columns, we may index $u^i_t$ as a tuple $(i,~t)$ and order these tuples in lexicographic order. 
The matrix $M$ is called the potential outcome matrix in the causal inference literature~\cite{rubin2005causal}.

\sloppy At the $t^\text{th}$ step of the $i^{\text{th}}$ trajectory, we observe 
$m^i_t = \mathcal{F}_{\text{trace}}(a^i_t, u^i_t)$, which is the entry in $M$ in the row corresponding to 
$a^i_t$ and the column corresponding to $u^i_t$. The counterfactual quantities of interest, 
$\tilde{m}^i_t = \mathcal{F}_{\text{trace}}(\tilde{a}^i_t, u^i_t)$ for $\tilde{a}^i_t \neq a^i_t$, are the missing entries in $M$ in the same column. In summary, we observe one entry per column of the matrix 
$M$ and we wish to estimate the missing values in the matrix. 

The task of filling missing values in a matrix based on its partially observed entries
is known as {\em Matrix Completion}~\cite{recht_completion}, a topic that has seen tremendous progress 
in the past two decades \cite{amazon_rec,tao_completion,cai_completion}.
However, standard matrix completion methods do not apply to our problem (see \S\ref{subsec:MC_discussion} for details).

We use a distributional invariance property of data collected using an \gls{rct} to complete the potential outcome matrix $M$. 
The key observation is that, in an RCT, the latent factors for trajectories collected under each of the policies will have the same distribution. For example, in Puffer's RCT, incoming users are assigned to an ABR algorithm at random. Therefore each ABR algorithm will ``experience'' the same distribution of underlying latent network conditions, which is precisely why we can compare their performance in the RCT. The same property helps us recover the matrix $M$, as we show next.

\subsection{Exploiting RCT for Matrix Completion}
\label{sub:MC+invar}
We use a minimal non-trivial example to give intuition about how we can exploit an \gls{rct} for matrix completion, before stating our main theoretical result. 

Consider a simple example where $A=2$ and $U=2n$, and the rank of potential outcome matrix $M$ is equal to 1.
Rank 1 implies that $M = au^T$ for some $a \in \mathbb{R}^2$
and $u \in \mathbb{R}^{2n}$ with $M_{\alpha, \beta} = a_{\alpha} \cdot u_{\beta}$.\footnote{Note that for readability, we are abusing notation by overloading $a$ and $u$ to refer to both the action and latent, and their encodings in the factorization.} 
Suppose we have $K=2$ policies, where each policy always chooses only one of the two actions. 
Furthermore, we consider an \gls{rct} setting.
That is, the distribution of latent factors across trajectories assigned to both policies
should be the same.

Without loss of generality, we can re-order the columns of $M$ so that the first $n$ columns correspond to the latent factors of the trajectories 
assigned to policy 1, and the second $n$ columns are those assigned to policy 2. Then the observed entries of matrix $M$ appear as 
\begin{equation*}
\begin{bmatrix}
{M}_{1,1} & {M}_{1,2} &\ldots  &{M}_{1,n} & \star & \ldots & \star & \star \\
\star & \star & \ldots & \star & {M}_{2,n+1} & \ldots & {M}_{2,2n-1} & {M}_{1,2n}\\
\end{bmatrix}
\end{equation*}
where $\star$ represents the missing values.

Let us consider recovering the missing observation $M_{2,1}$. 
For column $1$,  we know the observation under the first action, i.e. $M_{1,1}$. 
Due to rank $1$ structure, we have
\begin{equation}
\label{eqn:ratio}
\frac{{M}_{2,1}}{{M}_{1,1}} ~=~ \frac{a_2 \, u_1}{a_1 \, u_1} ~=~ \frac{a_2}{a_1}. 
\end{equation}
Therefore, to find $M_{2,1}$ (and by a similar argument, to find all missing entries of $M$), we need to estimate the ratio $\frac{a_2}{a_1}$. 

Due to the distributional invariance induced by \gls{rct}, the samples $u_1,\dots, u_n$ (which correspond to the latent factors encountered by policy 1) come from the same distribution as the samples $u_{n+1},\dots, u_{2n}$ (which correspond to the latent factors encountered by policy 2), for large enough $n$. Thus, their expected value should be equal:
\begin{equation}
\label{eqn:latent_sum}
    \frac{1}{n}\sum_{\beta=1}^{n} u_{\beta} \approx \frac{1}{n} \sum_{\beta=n+1}^{2n} u_{\beta} 
\end{equation}
\Cref{eqn:latent_sum} implies 
\begin{align}
\frac{\sum_{\beta=1}^n M_{1,\beta}}{\sum_{\beta=n+1}^{2n} M_{2,\beta}} & = 
\frac{\sum_{\beta=1}^n a_1 \cdot u_\beta}{\sum_{\beta=n+1}^{2n} a_2 \cdot u_\beta} \\
\approx \frac{a_1}{a_2}.
\end{align}
This provides precisely the quantity of interest in \Cref{eqn:ratio} based on the observed entries, enabling us to complete the matrix. 

\noindent \textbf{Formal Result.} This simple illustrative example relied on a convenient observational pattern (based on policies that always choose one action) and rank $1$ structure.
But the idea can be generalized.
If the trace includes D measurements, $M_{\alpha,\beta,\gamma} \in \mathbb{R}^{A\times U \times D}$ becomes a tensor rather than a matrix, where $\alpha$, $\beta$, and $\gamma$ index the actions, latent factors, and measurements, respectively.
The following theorem provides conditions where completion is possible for a rank $r$ tensor.
For more details and the proof, refer to \Cref{app:completion}.
\begin{theorem} \label{thm:1}
We can recover all entries of $M$ by only observing one $D-\text{dimensional}$ element in each column (corresponding to one latent and action) if the following is satisfied:
\begin{CompactEnumerate}
\item \textbf{(Low-Rank Factorization)} $M$ is a low-rank tensor ($\text{rank} = r$), i.e., it admits the following factorization: $M_{\alpha,\beta,\gamma} = \sum_{\ell=1}^r a_{\alpha \ell} u_{\beta \ell} z_{\gamma \ell}$.
\item \textbf{(Invertibility)} The factorization implies existence of a linear mapping from latent encoding to trace for each action. This linear mapping is invertible.
\item \textbf{(Sufficient measurements)} $D \ge r$.
\item \textbf{(Sufficient, Diverse Policies)} The number of policies $K \ge Ar$, and the matrix $\mathbf{S} \in \mathbb{R}^{Ar\times K}$ is full-rank where $\mathbf{S}_{w.D:(w+1).D,x} = \mathbb{E}[m|\text{action\_index}=w,\text{policy\_index} = x]\mathbb{P}(\text{action\_index}=w|\text{policy\_index}=x)$.
Linear independence of columns of $\mathbf{S}$ can be interpreted as diversity among policies (\Cref{app:completion}).
\end{CompactEnumerate}
\end{theorem}

\subsection{Discussion}
\label{subsec:MC_discussion}

\noindent{\bf Why not standard tensor completion?} 
Tensor completion methods~\cite{TC1,TC2,TC3,TC4} make several assumptions. 
First, the tensor $M$ must be (approximately) low rank, which \CausalSim also requires.
Low-rank structure holds in many real-world problems~\cite{udell2019big} and has been observed in network measurements, e.g.,  in traffic matrices~\cite{traffic1,traffic2,traffic3,traffic4} and network distance (i.e., RTT)~\cite{distance1,distance2,distance3}.  As an example of how it emerges in the problems we study in this paper, we use a simple model of congestion control in \Cref{app:synth:low_rank} to provide intuition about low-rank structure in ABR data. 

Second, the pattern of missing entries should be {\em random}. If the missing patterns is not random and depends on latent factors or the entries themselves~\cite{causal-matrix-completion}, standard approaches have difficulty recovering the tensor. This assumption does not hold in trace-driven simulation. Revealed entries are determined by the actions taken by the policies, which often use recent observations to make their decisions (e.g., an ABR policy may use recent throughput measurements). Hence the revealed/missing entries in a column are not random and depend on the entries in previous columns.

Third, a sufficient number of entries need to be revealed. For example, when $D=1$ (i.e., when $M$ is a matrix), the information theoretic lower bound to on the number of revealed entries needed to recover $M$ is  $4Ur-r^2$~\cite{xu2018minimal,observation-lower-bound}. Thus even for rank $r = 1$, it requires 4 entries per column, whereas only one entry per column is revealed in trace-driven simulation.

Since the second and third assumptions do not necessarily hold in our setup, we cannot use {\em existing} tensor completion methods. However, as we argued in \S\ref{sub:MC+invar}, exploiting the additional problem structure imposed by RCT data can make tensor completion feasible in certain conditions. 

\noindent{\bf Limitations of Theorem 4.1.} The proof of Theorem 4.1 (\Cref{app:completion}) provides an analytical method for recovering the tensor $M$ that generalizes the procedure described for the simple example in \S\ref{sub:MC+invar}. While this provides a theoretical basis for why tensor recovery is possible, the analytical approach is not practical. First, it relies on $M$ being exactly rank $r$; if it is approximately rank $r$, we have found the calculation to be brittle. Second, it  applies only to discrete action spaces. Third, it gives sufficient conditions for recovery, but they're not all necessary.  One reason is that the analytical method uses only {\em mean} invariance, i.e. the fact that the mean of the latent factors is the same across all policies (as in Eq.~\eqref{eqn:latent_sum}), even though RCT data has the stronger property that the entire {\em distribution} of latents does not depend on the policy. In the next section, we describe our practical implementation of \CausalSim that uses learning techniques and \glspl{nn} to overcome these limitations (at the expense of theoretical guarantees). 
\vspace{-2mm}

\section{\CausalSim: Algorithm}
\label{sec:solution}
\begin{figure}
    \centering
    \includegraphics[trim=16cm 10cm 19cm 7cm, clip, width=\linewidth]{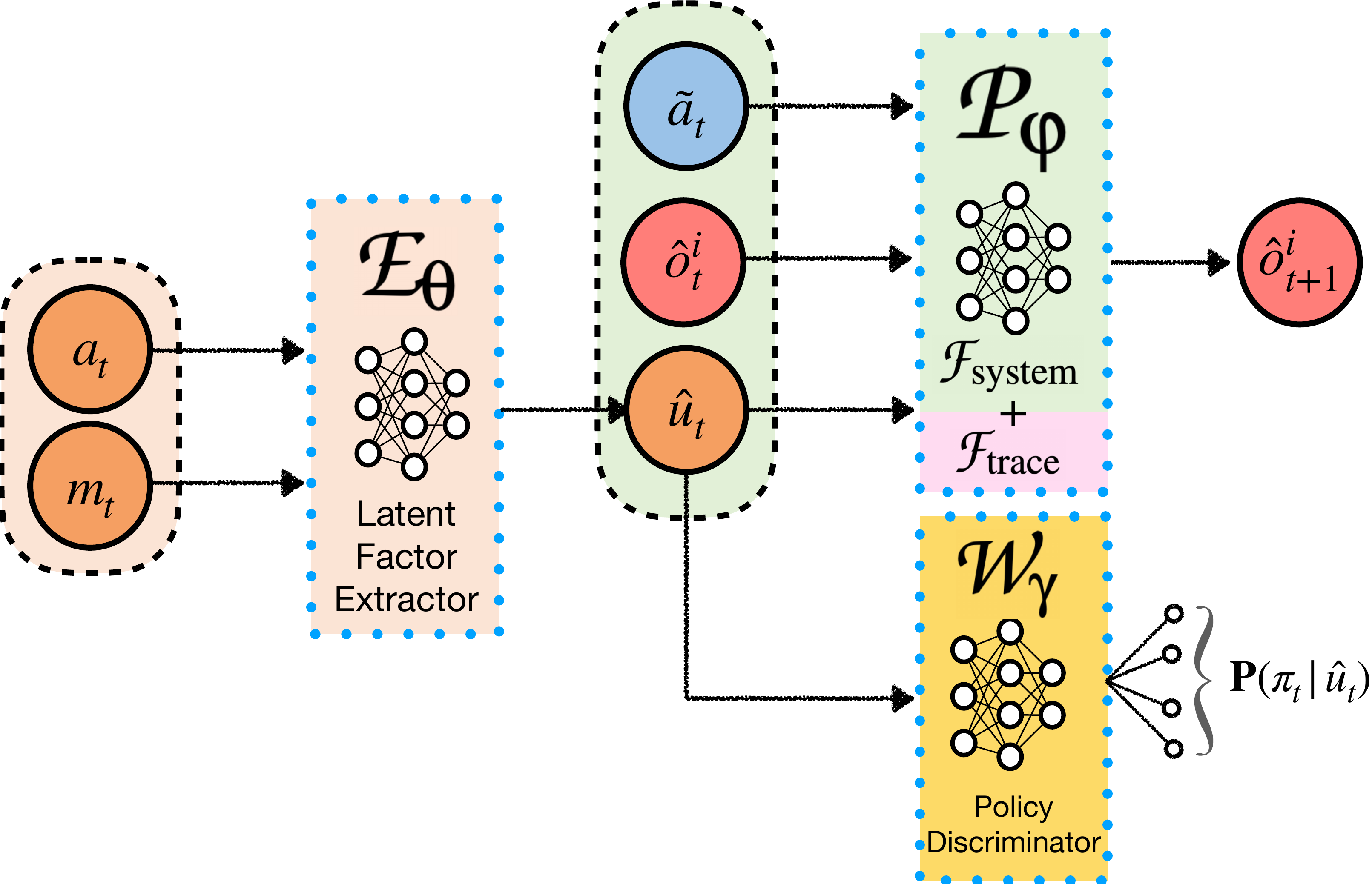}
    \caption{\CausalSim Architecture}
    \label{fig:overview_comb}
    \vspace{-2mm}
\end{figure}

\CausalSim builds upon the insights presented earlier but replaces the factorized model with a learning algorithm based on \glspl{nn}. For ease of notation,  we will drop the trajectory index for all variables in the dataset, e.g. we will refer to the latent factor $u_t^i: t\leq H_i, i \leq N$  as $u_t : t\leq H$.

\smallskip
\noindent {\bf \CausalSim architecture.} 
As discussed, \CausalSim aims to extract $u_t$ and learn 
$\mathcal{F}_{\text{trace}}$
and $\mathcal{F}_{\text{system}}$ from observed trajectories $(o_{t+1},o_t, m_t, a_t): t < H$. \Cref{fig:overview_comb} summarizes \CausalSim's algorithmic structure.

To extract latent factors, we use a \gls{nn} that takes in $a_t$ and $m_t$, and computes $\hat{u}_t$ (an estimate of $u_t$). To apply invariance on the extracted latents, i.e. distribution of $u_t$ being the same regardless of the policy applied to it, we use a \gls{nn} called the {\em Policy Discriminator}. This \gls{nn} aims to predict the policy pertaining to that sample given $\hat{u}_t$, and if invariance is upheld, it will fail to do so. Unlike the analytical approach, the policy discriminator can enforce policy invariance on the entire latent distribution, potentially improving the accuracy of the estimate. 

To calculate the counterfactual traces and observations, we need to learn $\mathcal{F}_{\text{trace}}$ and $\mathcal{F}_{\text{system}}$. However, we can simplify the learning problem by merging these two into one single combined function. Thus, we use a \gls{nn} that takes in counterfactual actions $\tilde{a}_t$, observation $o_t$ and estimated latent $\hat{u}_t$, and computes counterfactual observation $\tilde{o}_{t+1}$. Of course, we can explicitly use separate \glspl{nn} for $\mathcal{F}_{\text{trace}}$ and $\mathcal{F}_{\text{system}}$ if we require access to the simulated trace ($\tilde{m}_t$) values.

Overall, \CausalSim uses three \glspl{nn} for counterfactual simulation; $\mathcal{E}_{\theta}$ as the \emph{latent factor extractor}, $W_\gamma$ as the \emph{policy discriminator} and $\mathcal{P}_{\varphi}$ as the combination of $\mathcal{F}_{\text{trace}}$ and $\mathcal{F}_{\text{system}}$. \Cref{fig:overview_comb} depicts the structure. Training these \glspl{nn} is quick; on an A100 Nvidia GPU, \CausalSim's time to convergence on $56M$ data points ($230K$ streams) was less than $10$ minutes, and each simulation step in inference (on CPU) takes less than $150\mu s$. A full inference run  on the same volume of data takes less than 6 hours on a single CPU core and less than 20 minutes on 32 cores.

\smallskip
\noindent {\bf Training procedure.} 
\CausalSim's training procedure alternates between:
(i) training the policy discriminator using a discrimination loss $\mathcal{L}_\text{disc}$; and (ii) training other modules using an aggregated loss $\mathcal{L}_{\text{total}}$. \Cref{algo:CausalSim_train} provides a detailed pseudo code of this training procedure.

\begin{algorithm}[t]
\caption{\CausalSim Training}
\label{algo:CausalSim_train}
\begin{tikzpicture}
\node[draw=none, inner sep=0] (alg) {
\begin{varwidth}{\linewidth}
\algsetup{indent=24pt}
\begin{algorithmic}[1]
\STATE initialize parameter vectors $\gamma, \theta, \varphi$
\STATE initialize hyper-parameters $num\_disc\_it$, $\kappa$
\STATE initialize dataset $D \gets \{(o_i, m_i, a_i, \pi_i)\}_{i=1}^m$ from an RCT
\FOR{each iteration}
    \FOR{$num\_disc\_it$}
        \STATE sample minibatch $B \gets \{(o_l, m_l, a_l, \pi_l)\}_{l=1}^{b}$
        \STATE ${u}_l \gets \mathcal{E}_{\theta}(m_l, a_{l})$ for $l \in \{1, ... b\}$
        \STATE $\mathcal{L}_{\text{disc}} \gets \frac{1}{b}\Sigma_{l=1}^b \left[-\log \mathcal{W}_{\gamma}(\pi_l|{u}_l)\right]$
        \STATE $\gamma = \gamma - \lambda_{\gamma} \cdot \nabla_{\gamma} \mathcal{L}_{\text{disc}}$
    \ENDFOR
    \STATE sample minibatch $B \gets \{(o_{l+1}, o_l, m_l, a_l, \pi_l)\}_{l=1}^{b}$
    \STATE ${u}_l \gets \mathcal{E}_{\theta}(m_l, a_l)$ for $l \in \{1, ... b\}$
    \STATE $\mathcal{L}_{\text{disc}} \gets \frac{1}{b}\Sigma_{l=1}^b \left[-\log \mathcal{W}_{\gamma}(\pi_l|{u}_l)\right]$
    \STATE $\mathcal{L}_{\text{pred}} \gets \frac{1}{b}\Sigma_{l=1}^b \left[\left( o_{l+1}-\mathcal{P}_{\varphi}(o_l, a_l, u_l)\right)^2\right]$
    \STATE $\mathcal{L}_{\text{total}} \gets \mathcal{L}_{\text{pred}}-\kappa \cdot \mathcal{L}_{\text{disc}} $
    \STATE $\theta = \theta - \lambda_{\theta} \cdot \nabla_{\theta}\mathcal{L}_{\text{total}}$
    \STATE $\varphi = \varphi - \lambda_{\varphi} \cdot \nabla_{\varphi}\mathcal{L}_{\text{pred}}$
    
\ENDFOR
\vspace{0.6mm}
\end{algorithmic}
\end{varwidth}
};
    \node[inner sep=0] at (-3.4, -0.3) (nodeA) {};
    \node[inner sep=0] at (-3.4, 2.1) (nodeB) {};
    \draw[<->, color=red!70, line width=1pt] (nodeA) -- (nodeB) node [midway, sloped,font=\footnotesize, color=red!70, fill=white, inner sep=0] (TextNode) {Discriminator};
    
    \node[inner sep=0] at (-3.4, -3.5) (nodeC) {};
    \node[inner sep=0] at (-3.4, -0.5) (nodeD) {};
    \draw[<->, color=blue!70, line width=1pt] (nodeC) -- (nodeD) node [midway, sloped,font=\footnotesize, color=blue!70, fill=white, inner sep=0] (TextNode) {Simulation Modules};
\end{tikzpicture}
\end{algorithm}

\smallskip
\noindent {\bf Training the policy discriminator (Lines 5--10 in \Cref{algo:CausalSim_train}).} Distributional invariance means restricting the distribution of latent factors $u$ to be identical across policies. To that end, we first use $\mathcal{E}_{\theta}$ to extract latents $\hat{u}_t$, and then search for invariance violations via a {\em discriminator} \gls{nn}, a standard approach in the paradigm of adversarial learning~\cite{gan,ada}.
Specifically, the policy discriminator aims to predict the policy $\pi_i$ that took action $a_t$ from the estimated latent factor $\hat{u}_t$ (see \Cref{fig:overview_comb}).
Towards that, we use a cross-entropy loss to train the policy discriminator:
\begin{equation}
\label{eqn:CE}
    {\mathcal{L}_\text{disc}} = \mathbb{E}_{B}[-\log\mathcal{W}_{\gamma}(\pi|\hat{u})],
\end{equation}
where the expectation is over the a sampled minibatch $B$ from dataset $D$. 
We train the policy discriminator to minimize this loss, by repeating gradient decent $num\_disc\_it$ times, as the policy discriminator needs multiple iterations to catch up to changes in the latent factors.

\smallskip
\noindent {\bf Training simulation modules (Lines 11--17 in \Cref{algo:CausalSim_train}).} In this step, we need to impose consistency with observations, all while preserving the distributional invariance. Thus, we compute latent factors $\hat{u}_t$ with $\mathcal{E}_{\theta}$ and simulate the next step of the trajectory $\hat{o}_{t+1}$ with $\mathcal{P}_{\varphi}$. We use an aggregated loss to enforce consistency and invariance. This loss combines the negated discriminator loss with a quadratic consistency loss using a mixing hyper-parameter $\kappa$.
\begin{equation}
\label{eqn:total_loss}
    {\mathcal{L}}_{\text{total}} = \mathbb{E}_B \left[\left(o_{t+1} - \hat{o}_{t+1} \right)^2 \right]   - \kappa \mathcal{L}_\text{disc},
\end{equation}
where the expectation is over the a sampled minibatch $B$ from dataset $D$. 
Here, we used a quadratic loss function, but one could use any consistency loss fit to the specific type of variable (e.g. Huber loss, Cross entropy, ...).

Note the negative sign of discriminator loss, which means we train these NNs to maximize discriminator loss i.e., fool the discriminator to ensure policy invariance. 
If the extracted latent factors are policy invariant, the policy discriminator should do no better at its task than guessing at random.

\smallskip
\noindent {\bf Counterfactual estimation.} To produce counterfactual estimates, as described above, 
the estimated latents $\hat{u}_t$ are extracted from observed data. 
Using the extracted latents factors, along with the learned combined function $\mathcal{P}_\gamma$, we start with 
$o_{1}$ and predict counterfactual observations $\hat{o}_{t+1}$, one step at a time.
\vspace{-1mm}
\section{Evaluation}\label{sec:results}

We evaluate \CausalSim's ability to do accurate counterfactual simulation (\S\ref{sec:real_abr} and \S\ref{subsec:buffer_eval}) using trace data from one real-world and one synthetic dataset. As a rigorous proof of concept, we debug and improve an ill-performing ABR policy with \CausalSim (\S\ref{subsec:puffer_bayes}), and verify it through deployment on a public ABR testing infrastructure.
Our baselines are as follows:
\begin{CompactEnumerate}
    \item \emph{ExpertSim}: Uses the analytical model described in  \S\ref{subsub:expert_sim}.
    \item \emph{SLSim}: Uses a standard supervised-learning technique to learn system dynamics from data, as described in  \S\ref{subsub:data-driven}.
\end{CompactEnumerate}

Finally, we show how \CausalSim enables trace-driven simulation in problems where defining an \emph{exogenous trace} is not straightforward and traditional trace-driven simulation is not applicable (\S\ref{subsec:lb}). 
Further supporting experiments in the appendix provide more details about how \CausalSim operates (\S\ref{app:puffer:all}, \S\ref{app:puffer_confusion}, \S\ref{app:sub:difficulty}, \S\ref{sec:fine_grained}, \S\ref{sec:puffer_hyper_params}, \S\ref{app:subsec:puffer_separated}, \S\ref{app:abr_synth:extended_eval}, \S\ref{app:subsec:abr_rl}, \S\ref{app:synth:low_rank} and \S\ref{app:lb_lat_match}). 


\subsection{Simulation Accuracy}\label{sec:real_abr}

We use \CausalSim to predict the end performance of \gls{abr} policies, and compare them with ground truth data. We explore the same two metrics reported by Puffer to evaluate algorithms; 1) stall rate, which is the fraction of time a user spent rebuffering, i.e. paused and waiting for a new chunk to download; 2) average Structural Similarity Index Measure (SSIM) in decibels, which is a perceptual quality metric. Our ground truth data comes from public logs of `slow streams' on Puffer. Whenever a client initiates a video streaming session in Puffer's website, a random \gls{abr} algorithm is chosen and assigned to that session. Sessions are logged (buffer levels, chunk sizes, timestamps, download times, etc) anonymously and the data is available for public use. Our dataset contains more than $230K$ trajectories from an \gls{rct} during July 2020 to June 2021, where five \gls{abr} algorithms 
(\gls{bba}, BOLA1, BOLA2, Fugu-CL, Fugu-2019) 
were evaluated. 
Exhaustive details of the setup and data can be found in \S\ref{app:puffer:alg_data}.

\subsubsection{Can \CausalSim simulate a policy it has not seen?}
\label{subsubsec:results:puffer:can_sim}
We choose one of \gls{bba}, BOLA1, and BOLA2\footnote{We exclude Fugu as a test policy since we could not reproduce its logged actions (see \S\ref{app:puffer:alg_data})}. as the new policy that we want to simulate, and call it the \emph{target} policy.
The remaining four policies are called \emph{source} policies.
Traces assigned to the four source policies comprise our training dataset, which we use for training \CausalSim and the two baselines.
The goal is to simulate the outcome of applying the target policy on trajectories assigned to any of the source policies. 

\begin{figure}
    \centering
    \begin{subfigure}{\linewidth}
        \centering
        \includegraphics[width=\linewidth]{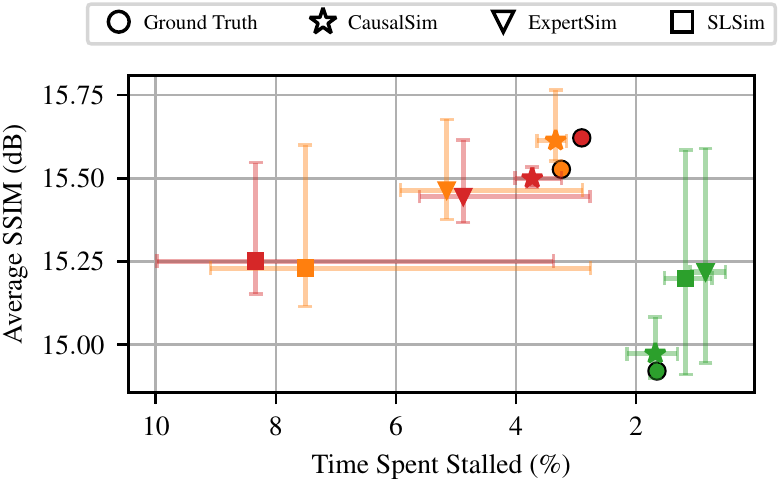}
        \caption{}
        \label{fig:full_scatter}
    \end{subfigure}
    \begin{subfigure}{\linewidth}
        \centering
        \includegraphics[width=\linewidth]{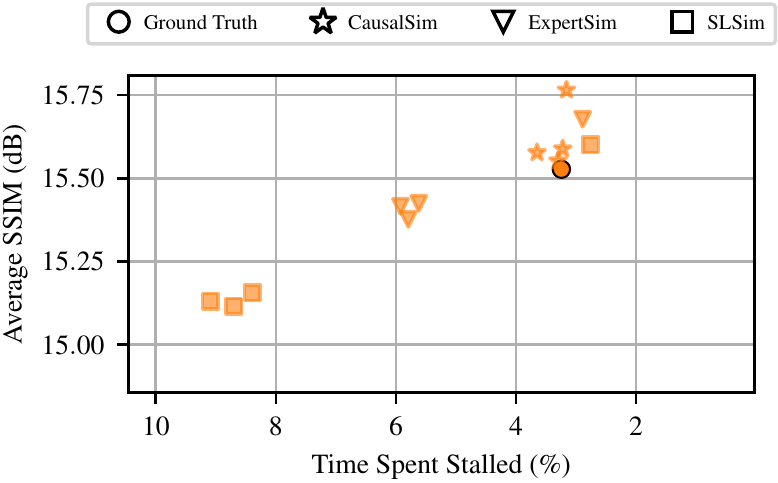}
        \caption{}
        \label{fig:bola1_partial_scatter}
  \end{subfigure}
    \caption{\textbf{(a)} In a real-world dataset of live video streaming, \CausalSim is the most faithful, compared to traditional trace-driven (ExpertSim) or data-driven (SLSim) simulators. Colors indicate different \textbf{target} ABR algorithms. \textbf{(b)} Predictions for BOLA1, separated by the source policy. Each point indicates a different \textbf{source} ABR algorithm. ExpertSim and SLSim predictions carry over biases of the source data, while \CausalSim mitigates the bias.}
    \label{fig:scatter_plots}
\end{figure}

\Cref{fig:full_scatter} plots the stall rate and SSIM in the simulated trajectories and ground truth, denoting each target policy with a different color. Four source policies give us four separate predictions per target policy and simulator. Each point depicts the average of these four predictions, and the intervals show the minimum and maximum among the four.
For either metric, \CausalSim is the most faithful to ground truth among all simulators. For instance, in stall rate, \CausalSim's relative error spans $2-28\%$, while ExpertSim spans $49-68\%$ and SLSim spans $29-187\%$. 
CausalSim may not always predict the correct relative ordering among policies with close performance. For example, BOLA1 and BOLA2 (shown in orange and red) have similar performance in both stall rate and SSIM. CausalSim predicts that these policies are similar but it infers their relative ordering incorrectly. However, CausalSim avoids the large errors made by the baseline simulators. In absolute terms, its predictions are close to the ground truth.

\CausalSim also has the most consistent predictions across different \textbf{source} policies, because it removes the biases of the source policies. As an example, we investigate all four simulation results for BOLA1 in \Cref{fig:bola1_partial_scatter}. SLSim and ExpertSim's simulation results are only good when the source algorithm is BOLA2 (a similar algorithm to BOLA1 performance-wise). However, their predictions are far off from the ground truth for the other three source algorithms. \CausalSim's simulation results, on the other hand, are all close to the ground truth target. Appendix \S\ref{app:subsec:puffer_separated} demonstrates the same observation for other target algorithms, i.e. BBA and BOLA2.

\subsection{Case Study: \CausalSim in the Wild}\label{subsec:puffer_bayes}
\label{subsec:causalsim-casestudy}
An accurate simulator allows researchers to debug and improve protocols without repeated and invasive deployments. We shall demonstrate this with \CausalSim, by improving a well-known ABR policy, and verifying our findings with a real-world deployment on Puffer.

\begin{figure}
    \centering
    \includegraphics[width=\linewidth]{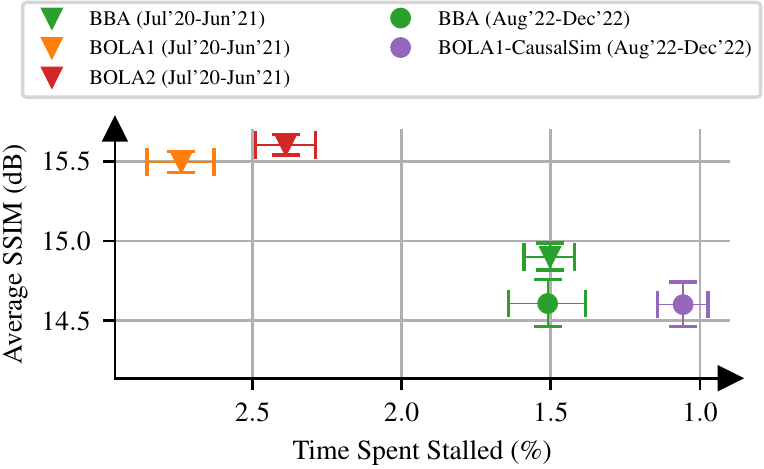}
    \caption{In an experiment preceding this work, BOLA1 exhibits high stalling. By deploying a BOLA1 variant in a later experiment \textbf{\CausalSim improved the stall rate by $2.6\times$}, with comparable quality to \gls{bba}. User population is `slow streams' and error bars denote 2.5\%--97.5\% confidence intervals.}
    \label{fig:puffer_bayes:rct}
\end{figure}

Recall that in the particular RCT we used in \S\ref{sec:real_abr}, five \gls{abr} algorithms (\gls{bba}, BOLA1, BOLA2, Fugu-CL, Fugu-2019) were evaluated. \Cref{fig:puffer_bayes:rct} shows the result of this evaluation for \gls{bba}, BOLA1 and BOLA2, across `slow streams'.\footnote{The data for this plot comes directly from Puffer~\cite{puffer_rct1, puffer_rct2}.} Similar to \Cref{fig:full_scatter}, the X-axis shows the stall rate, and the Y-axis is the average SSIM. BOLA1 exhibited 82\% more rebuffering compared to \gls{bba}. A revised version of BOLA1, called BOLA2, was deployed alongside it, since the Puffer team and the authors of BOLA believed the SSIM metric (in decibels) is incompatible with the protocol~\cite{marx2020implementing}. This new version had 12.8\% less rebuffering and slightly higher quality, but still far too much stalling compared to \gls{bba}.

\begin{figure}
    \centering
    \includegraphics[width=\linewidth]{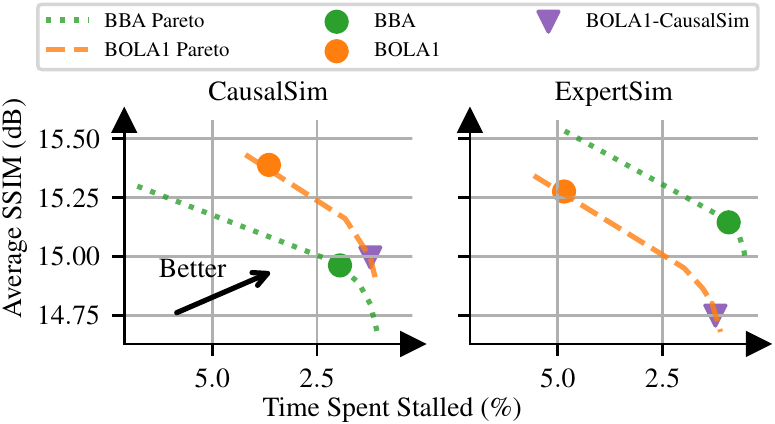}
    \caption{Pareto frontier curves for BOLA1 and \gls{bba} variants. \textbf{CausalSim correctly predicts BOLA1's potential}, while ExpertSim fails to do so.}
    \label{fig:puffer_bayes:pareto}
    \vspace{-4mm}
\end{figure}

BOLA1 is an ABR policy with two hyperparameters, similar to \gls{bba}, and our hypothesis was that BOLA1 uses sub-optimal hyperparameters.
To investigate this, we used the logged data pertaining to that plot along with \CausalSim to exhaustively analyze the performance of BOLA1 and \gls{bba} for a range of hyperparameters. Using Bayesian Optimization\footnote{We use a Gaussian Process prior with a Matern Kernel~\cite{matern2013spatial}.}, we explored the parameter space and created a Pareto frontier curve for each policy.  During this process, we evaluated over 150 different algorithms in two days, which is achievable only in a simulator.
Each curve demonstrates the trade-off between quality and stall rate in that policy. \Cref{fig:puffer_bayes:pareto} presents the curves, where the left and right plots show CausalSim and ExpertSim predictions. For ease of comparison, we highlight where the original BOLA1 and \gls{bba} lie. CausalSim confirms our suspicion; the curve for BOLA1 is strictly better than that of \gls{bba}. We can revise the hyperparameters in BOLA1 for an improved BOLA1 variant, henceforth called `BOLA1-CausalSim'. We chose BOLA1-CausalSim, such that it would have better stall rate and marginally better SSIM compared to \gls{bba}.

Interestingly, ExpertSim predicts the complete opposite.
It predicts that not only will \gls{bba} always improve on any BOLA1 variant in at least one metric, but also that any BOLA1 variant will stall more. This serves as a great opportunity to test \CausalSim's edge compared to traditional (biased) trace-driven simulation, which is used in prior work~\cite{mpc,FESTIVE,pensieve}. The results of BOLA1-CausalSim's deployment can be seen in \Cref{fig:puffer_bayes:rct}. Considering confidence intervals, it is clear that it stalls less than \gls{bba}; in fact, \gls{bba} stalls 43\% more than BOLA1-CausalSim on average. The confidence intervals for quality are wide and will need more data to be separable\footnote{Updated plots can be found on the `Experimental Results' page of the Puffer website~\cite{puffer_daily}, under "Current experiment, full contiguous duration, slow streams only".}, but based on the ongoing trend, BOLA1-CausalSim will have similar quality compared to \gls{bba}. 

Our goal was to show \CausalSim's potential, and for that we targeted one of several plots on Puffer (`slow streams'). We could have chosen a different plot to optimize on, but it would not affect the takeaway. Note that our opportunities for deployment on Puffer are limited, as other researchers use Puffer as well; hence we only deployed one BOLA1 variant. Furthermore, we hoped to also compare \CausalSim's prediction of stall rate and quality with the deployment results, but the client and network population has clearly changed; as shown in \Cref{fig:puffer_bayes:rct}, \gls{bba} achieves a different SSIM value for the two periods of time. Since CausalSim's predictions are based on data from the previous RCT, directly comparing the predicted values to results from the new RCT isn't meaningful. However, as our results show, the old RCT data allows us to compare different schemes. For example, \CausalSim predicts BBA stalls 58\% more than BOLA1-CausalSim on network distribution of the old RCT, which is reasonably close to the 43\% observed in the new RCT (ignoring confidence intervals).

\subsection{A Closer Look at Simulated Trajectories}\label{subsec:buffer_eval}
For a deep dive in simulator accuracy, we focus on buffer occupancy level, a key indicator of ABR algorithm behavior. Ideally, we would like to compare simulated trajectories to ground truth. But this isn't possible using real trace data, since it requires us to have multiple traces of different policies running under the exact same underlying path conditions. 
To overcome this issue, we resort to distributional evaluation.
Puffer data is collected in an \gls{rct} setting; hence the characteristics of network paths assigned to each policy is the same.
If we accurately simulate the target policy on traces assigned to one of the source policies, the distribution of each variable (e.g. buffer level) must be similar in the simulated trajectory and ground truth trace assigned to the target policy.
This motivates using distributional similarity as our performance metric.

To quantify the similarity of two distributions, we use the \gls{emd}~\cite{rubner1998metric}.
We can calculate \gls{emd} for one-dimensional distributions as 
$
    \text{\gls{emd}}(\mathcal{P}, \mathcal{Q}) = \int_{-\infty}^{+\infty} |\mathcal{P}(x)-\mathcal{Q}(x)| dx,
$
where $\mathcal{P}$ and $\mathcal{Q}$ are the \gls{cdf}s of $p$ and $q$, respectively. 
A small \gls{emd} between two distributions implies that they are similar.

\begin{figure}
    \begin{subfigure}{0.49\linewidth}
        \centering
        \hspace{-2mm}
        \includegraphics[width=\linewidth]{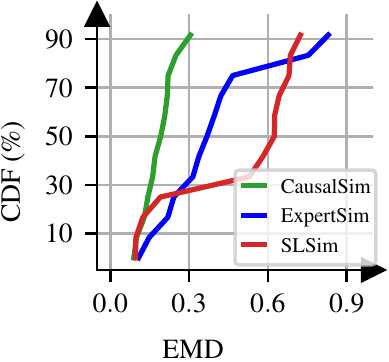}
        \caption{}
        \label{fig:puffer:emd_cdf}
    \end{subfigure}
    \begin{subfigure}{0.49\linewidth}
        \centering
        \hspace{-2mm}
        \includegraphics[width=\linewidth]{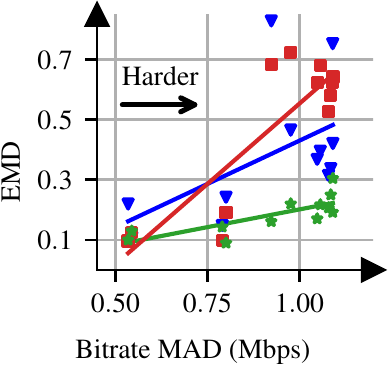}
        \caption{}
        \label{app:fig:puffer_difficulty_causalsim}
    \end{subfigure}
    \vspace{-3mm}
    \caption{On average, \CausalSim improves the EMD distance metric compared to ExpertSim and SLSim by 53\% and 61\% respectively. \textbf{(a)} Distribution of \CausalSim, ExpertSim, and SLSim EMDs over all possible source/target choices. \textbf{(b)} Error (\gls{emd}) increases for baseline as simulation scenarios get harder, but \CausalSim maintains good accuracy.}
    \label{fig:puffer:three_dists}
    \vspace{-2mm}
\end{figure}

\Cref{fig:puffer:emd_cdf} shows the CDF of the \gls{emd} (between actual and simulated buffer level distributions) for \CausalSim and baselines, over all possible source/target policy pairs. 
\Gls{emd} of \CausalSim is smaller than \gls{emd} of baselines across almost all experiments.
In terms of the average \gls{emd} across all experiments, \CausalSim bests ExpertSim and SLSim by 53\% and 61\% respectively.
\Cref{fig:motiv:buffer} visualized differences in buffer level distributions for the simulation scenario where BOLA2 and \gls{bba} are source and target policies, respectively.
To observe buffer level distributions for all scenarios, refer to \Cref{app:fig:puffer:all}.

In about 30\% of cases, SLSim is slightly better than \CausalSim. 
These cases are ``easy'' simulation scenarios where the source and target policies make similar actions (For more details see \S\ref{app:sub:difficulty}). 
In these cases, the \gls{emd} is low for both \CausalSim and baseline simulators ($< 0.15$), and all perform well.
For instance, \Cref{app:fig:puff:easy} (in the Appendix) shows source, target, and simulated buffer level distribution in an easy scenario, where BOLA2 and BOLA1 are the target and source policies respectively.
In this example, all simulated distributions match the target distribution quite well. 

\Cref{app:fig:puffer_difficulty_causalsim} shows where \CausalSim most shines, i.e. hard simulation scenarios.
The Y-axis is the error (\gls{emd}), and the X-axis is the mean absolute difference (MAD) between actions taken by the source policy and the target policy, in SLSim simulation.
The larger the action difference, the harder the scenario (\S\ref{app:sub:difficulty}).
As we move toward harder scenarios, the error increases significantly for the baselines, while \CausalSim is more robust.


\subsubsection{Additional experiments}
\label{subsub:puffer_additional_exp}
We perform further evaluations of \CausalSim in the ABR environment. 
Due to space constraints, we summarize these results here and defer details to the appendix. 

\noindent \textbf{A more fine-grained evaluation.} 
In the results above, we evaluated the performance of \CausalSim and baselines using the distribution of buffer occupancy across the whole population. 
One way to further validate the results is to test whether they will hold on carefully partitioned sub-populations. 
In \S\ref{sec:fine_grained}, we show that this is indeed the case when the sub-populations are partitioned according to the Min \gls{rtt}, a network property that is independent of the selected \gls{abr} algorithm in Puffer. 

\noindent \textbf{Hyperparameters tuning.} Counterfactual estimation (\S\ref{sec:sim-is-ce}) is inherently an \gls{ood} prediction task. 
Hence, typical supervised-learning hyper-parameter tuning methods do not work. 
In \S\ref{sec:puffer_hyper_params}, we describe and evaluate \CausalSim's hyper-parameter tuning procedure.

\noindent \textbf{Ground truth evaluation.}
Real data never comes with ground truth counterfactual labels.
As a result, we cannot evaluate \CausalSim's simulations for each time step in real data, but we can do this in a reproducible synthetic environment. 
In \S\ref{app:abr_synth:extended_eval}, we evaluate \CausalSim using ground truth counterfactual labels and show that it still outperforms baselines in the \gls{mape} metric.\footnote{Let $\mathbf{\hat{p}}=\{\hat{p}_i\}_{i=1}^N$ and $\mathbf{p}=\{p_i\}_{i=1}^N$ denotes the vectors of predicted and ground truth quantity of interest, respectively. Then, \gls{mape} is defined as 
$
    \text{MAPE}(\mathbf{p}, \mathbf{\hat{p}}) = \frac{100}{N}\sum_{i=1}^N \frac{\left|\hat{p}_i - p_i\right|}{p_i}
$.} Specfically, \CausalSim achieves an \gls{mape} of (${\sim}5\%$), which is significantly lower than both ExpertSim's and SLSim's (${\sim}10\%$).


\subsection{A Second Example: Server Load Balancing}
\label{subsec:lb}

We now focus on simulating load balancing policies with heterogeneous servers, where defining an exogenous trace is not possible and therefore standard trace-driven simulation is not applicable. 
This example shows how \CausalSim opens up new avenues in trace-driven simulation.

We use a synthetic environment which consists of $N=8$ servers (and a queue for each) with different processing powers, a load balancer, and a series of jobs that need to be processed on these servers. 
Each job has a specific size which is unknown to the load balancer.
Each server can process jobs at a specific rate $\{r_i\}_{i=1}^N$, which is also unknown to the load balancer. 
The load balancer receives jobs and must assign them to one of $N$ servers. 
Assuming the $k^{\text{th}}$ arriving job has size $S_{k}$ and gets assigned to server $a_k$, the job processing time will be $S_k / r_{a_k}$.
If this job is not blocked by some other job being processed, its latency will equal its processing time.
If it is blocked, and the jobs ahead of it in the queue take $T_k$ to be processed, the incurred latency is $S_k / r_{a_k} + T_k$. 

We generate a collection of 5000 trajectories each with 1000 steps and use 16 policies in the load balancer. For a detailed explanation of the policies, job size generation process, and server processing rates,  refer to \S\ref{app:lb:alg_data}.

\vspace{-2mm}
\subsubsection{Experiment setup}\label{lb:exp_setup}
The aim of this experiment is to evaluate whether we can simulate new unseen server assignment policies in this environment, using traces collected with other policies.
Recall that while we observe the processing time of each job, the actual size of the job is not observed, i.e., it acts as the latent factor in this problem. 
For all simulators, we assume access to $\mathcal{F}_{\text{system}}$ (the queue model) and focus on the more challenging task of learning $\mathcal{F}_{\text{trace}}$ and estimating the counterfactual traces $\hat{m}_i^t$ for $i \leq 5000$, and $t \leq 1000$. 
Algorithmically, this translates to enforcing consistency for the observed traces ($m_t$), rather than the observations ($o_t$) (see \S\ref{sec:solution}). 
The trace we collect is the processing time when using a \emph{source} server assignment policy. 
To simulate a \emph{target} server assignment policy, we need to estimate the processing time of a job on servers other than the one where its processing time was measured (without knowing either the job size or the server processing rates).

Standard trace-driven simulation assumes an exogenous trace (job processing time), but this is the same as assuming servers have equal processing rates. This contradicts the problem setup, and standard trace-driven simulation (analogous to ExpertSim in ABR) is not applicable to this problem. Thus, we compare \CausalSim with \emph{SLSim} simulations.
SLSim (realized by an \gls{nn}) takes as input the observed processing time and the target server, and its output is the processing time under the targeted server. 
However, the observed and target processing time are always the same in training data, and hence it is impossible for SLSim to learn the true dynamics (e.g., the server's underlying processing power).
\CausalSim sidesteps this problem by explicitly estimating latent factors.   
For details regarding the network architecture and training details for both SLSim and \CausalSim, refer to \Cref{table:lb_train} in the appendix.

\noindent\textbf{Performance Metric.}
We compare \CausalSim and SLSim with the underlying ground truth using the \gls{mape} metric.

\subsubsection{Can \CausalSim Faithfully Simulate New Policies?}
As is done in the ABR case studies, we train \CausalSim and SLSim models based on a dataset generated using all policies except one, which will be the target policy. 
We use the same hyper-parameter tuning approaches explained in \S\ref{sec:puffer_hyper_params} for \CausalSim and \S\ref{sec:slsim_hyper_params} for SLSim.
We carry out this evaluation on eight target policies. 
We evaluate the performance for each pair of source-target policies, as was done in \S\ref{sec:real_abr}. 
In total, we have 120 different source/target policy pairs.

\begin{figure}
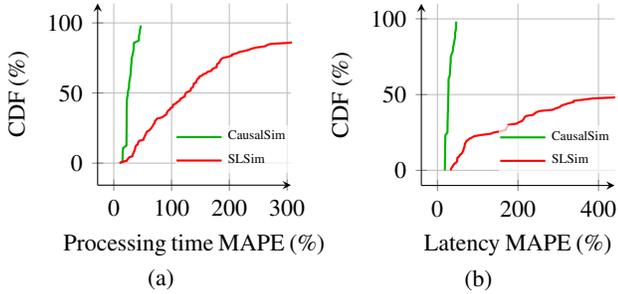

    \begin{subfigure}{0.49\linewidth}
        \centering
        \hspace*{-0.03\linewidth}
        \begin{tikzpicture}
            \begin{axis}[height=4cm,
                         width=\linewidth,
                         axis lines=left,
                         xlabel={Processing time MAPE (\%)},
                         ylabel={CDF (\%)},
                         font=\small,
                         legend columns=1,
                         enlargelimits=true,
                         legend style={at={(0.4,0.2)}, 
                              anchor=west, fill=white, fill opacity=0.7, draw opacity=1, 
                              text opacity=1, draw=none, font=\tiny, inner sep=0},
                              legend cell align={left},
                         xmin=-0.084609950601731, xmax=278.722135447377,
                         ymin=-4.91071428571429, ymax=103.125,
                         ylabel near ticks,
                         xlabel near ticks,
                         grid=both,
                         every x tick scale label/.style={
                        at={(1,0)},xshift=1pt,anchor=west,inner sep=0pt}
                         ]
                \input{figures/raw_data/load_balancing/ptimes_mape}
                \legend{ \CausalSim, SLSim };
            \end{axis}
        \end{tikzpicture}
        \vspace{-6.3mm}
        \caption{}
        \label{fig:lb:mape_cdf_ptimes}
        \vspace{2mm}
    \end{subfigure}
    \begin{subfigure}{0.49\linewidth}
        \centering
        \hspace*{-0.03\linewidth}
        \begin{tikzpicture}
            \begin{axis}[height=4cm,
                         width=\linewidth,
                         axis lines=left,
                         xlabel={Latency MAPE (\%)},
                         ylabel={CDF (\%)},
                         font=\small,
                         legend columns=1,
                         enlargelimits=true,
                         legend style={at={(0.4,0.2)}, 
                              anchor=west, fill=white, fill opacity=0.7, draw opacity=1, 
                              text opacity=1, draw=none, font=\tiny, inner sep=0},
                              legend cell align={left},
                         xmin=0, xmax=400,
                         ymax=100, 
                         ymin=0,
                         ylabel near ticks,
                         xlabel near ticks,
                         grid=both,
                         every x tick scale label/.style={ at={(1,0)},xshift=1pt,anchor=west,inner sep=0pt}
                         ]
                \input{figures/raw_data/load_balancing/latency_mape}
                \legend{ \CausalSim, SLSim };
            \end{axis}
        \end{tikzpicture}
        \vspace{-2mm}
        \caption{}
        \label{fig:lb:mape_cdf_latency}
        \vspace{2mm}
    \end{subfigure}
  \vspace{-2mm}
  \caption{Distribution of \CausalSim and SLSim MAPEs over all source target pairs.
  }
   \vspace{-4mm}
\end{figure}

In \Cref{fig:lb:mape_cdf_ptimes} and \Cref{fig:lb:mape_cdf_latency}, we show the CDF of the \gls{mape} of estimating the processing time and the latency, respectively,  using both \CausalSim and SLSim. 
As evident in these two figures, \CausalSim's error is significantly lower than that of SLSim for both the processing time and latency. 
In particular, the median MAPE when estimating processing time/latency is 24.4\%/27.0\% for \CausalSim  and 124.3\%/467.8\% for SLSim. For a complementary view, we compare the latent factors \CausalSim extracts to the real latent job sizes and observe how closely they match, in \S\ref{app:lb_lat_match} in the appendix.

\vspace{-1mm}
\section{Related-Work}
\label{sec:related-work}

\noindent\textbf{Data-driven simulation.}
Traditional packet-level simulators~\cite{opnet,ns3,mininet} tend to sacrifice either scalability or accuracy when simulating large networks.
MimicNet~\cite{mimicnet} and DeepQueueNet~\cite{deepQ} use machine learning to improve simulation speed of datacenter networks.
The aforementioned approaches are all full-system packet-level simulators, whereas \CausalSim focuses on trace-driven simulation of a specific system component and must therefore deal with latent factors and biases present in trace data.

A very recent work, Veritas~\cite{veritas} (published on arXiv in Aug. 2022), models trace-driven simulation for \gls{abr} as a \gls{hmm} with a known emission process. This is equivalent  to assuming that $\mathcal{F}_{\text{trace}}$ is known in our model (see Eq.~\eqref{eqn:emission}).
Veritas uses the Viterbi algorithm to decode the latent factors, which are then used for counterfactual simulation. 
\CausalSim solves a more general problem where $\mathcal{F}_{\text{trace}}$ is not known and must be learned. It therefore requires less knowledge of the system's latents and underlying dynamics to apply. On the other hand, \CausalSim requires RCT data whereas Veritas does not. Comparing the fidelity of these approaches using real-world ABR data would be interesting future work (Veritas evaluates its method in a network emulator).

Panthon's calibrated emulators~\cite{yan2018pantheon} model the end-to-end behaviour of a network path with a simple model including a handful of parameters, e.g., bottleneck link rate, constant propagation delay, etc., which are tuned to fit a collection of packet traces collected from this path using a variety of congestion control protocols.
iBox~\cite{ashok2022data} extends this approach by modeling cross-traffic. 
\CausalSim does not assume any known model for the dynamics of the network. Furthermore, it has access to only a single trace from each network path.

\noindent\textbf{Policy evaluation.}
Policy evaluation techniques such as Inverse Propensity Scoring~\cite{ips} and Doubly Robust~\cite{bartulovic2017biases}  aim to predict population-level performance statistics for a given intervention.  WISE~\cite{wise} builds a Causal Bayesian Network from the data that is able to answer interventional (what-if) queries about the future, but the method requires absence of latent confounding variables.  Sage~\cite{sage} uses a Causal Bayesian Network model with latent factors to diagnose performance issues in microservice applications. It answers what-if questions about how interventions like changing the resources allocated to a microservice impacts the end-to-end application latency. Trace-driven simulation is distinct from all these methods, in that it requires counterfactual predictions of how an intervention would have changed specific previously-measured trajectories rather than how it changes population-level statistics.\footnote{\Cref{app:related} provides a broader overview of the causal inference literature.}
\section{Concluding Remarks}
\label{sec:discussion}

The exogenous trace assumption is central to traditional trace-driven simulation.
\CausalSim relaxes this key assumption, by modeling the intervention effect on the trace and learning to replay the trace in an unbiased manner. We showed how this improves the accuracy of trace-driven simulation using real-world ABR data, and how \CausalSim provides insights for algorithm improvement that are in contrast with standard trace-driven simulators' predictions, which we validated in a real-world deployment. Furthermore, we showed how this expands the applicability of trace-driven simulation to problems where defining an exogenous trace is not possible by applying it to heterogeneous server load balancing. We believe \CausalSim could be applied to many other system simulation tasks.

\CausalSim opens up several interesting paths for future work. First, evaluating \CausalSim in problems with a higher-dimensional latent factors would be interesting. Second, it is a natural next step to use \CausalSim for more complex policy optimization methods, e.g., using reinforcement learning. Last, as discussed in \S\ref{subsec:MC_discussion}, our theoretical analysis of \CausalSim's approach, i.e. exploiting the policy invariance of latent factors distributions, is not tight, and improving it could potentially relax the assumptions of our analytical method.
\section{Acknowledgement}
\label{sec:ack}

We thank our shepherd Keith Winstein for in-depth suggestions, and our reviewers for insightful comments. We thank the Puffer team, specifically Emily Marx and Francis Y. Yan for providing us with the data we used in \S\ref{sec:real_abr} and the algorithm deployment in \S\ref{subsec:puffer_bayes}.
This work was supported by NSF grants 1751009 and 1955370, an award from the SystemsThatLearn@CSAIL program, and a gift from Intel as part of the MIT Data Systems and AI Lab (DSAIL).
A. Alomar and D. Shah were supported in part by DSO-Singapore project, MIT-IBM project on Causal representation learning and NSF FODSI project.

\bibliographystyle{plain}
\bibliography{paper}

\clearpage
\appendix
\begin{appendices}
\section{Tensor Completion \\with policy invariance}
\label{app:completion}
Here, we discuss a more generic version of the problem considered in \S\ref{sub:MC+invar} from the lens of tensor completion.
Specifically, in \S\ref{sec:MatrixCompletion}
we considered the simplified setting where the trace was considered to be one-dimensional. Here, 
we shall consider higher dimensional traces. This, naturally suggests using the lens of 
Tensor instead of Matrix completion. We will also discuss how higher dimensional trace can 
enable recovery of more complex system dynamics or models compared to the simple solution we discussed in 
\S\ref{sec:MatrixCompletion} for rank $1$ setup. 

\medskip
\noindent{\bf Potential Outcomes Tensor.} As considered in \S\ref{sec:MatrixCompletion} let all possible
actions be denoted as $[A] = \{1,\dots, A\}$ for some $A \geq 2$. Let the trace be of $D$ dimension. 
As before, we have $N$ trajectories of interest with trajectory $i \in [N]$ being
of length $H_i \geq 1$ time steps. As before, let $U = \sum_{i=1}^N H_i$. 

Consider an order-$3$ tensor $M$ of dimension $A \times U \times D$, where 
$M = [m_{\alpha \beta \gamma}: \alpha \in [A], \beta \in [U], \gamma \in [D]]$
with $m_{\alpha \beta \gamma}$ corresponds to the $\gamma$th co-ordinate of 
the $D$-dimensional trace corresponding to action $a_t = \alpha \in [A]$ 
when latent factor is $u_{i,t}$ with $\beta$ corresponding to enumeration
of $(i,t)$ for some $i \in [N]$ and $t \leq H_i$. Recall that, as explained
in Section \ref{sec:MatrixCompletion}, all possible $(i,t):  t \leq H_i, i \in [N]$ 
are mapped to an integer in $[U]$. We call this tensor $M$ as the Potential 
Outcomes Tensor. 

Indeed, if we know $M$ completely, then we can answer the task of simulation or
counterfactual estimation well since we will be able to estimate the mediator
for each trajectory under a given possible sequence of counterfactual actions, 
and subsequently estimate the counterfactual observation (assuming we could 
learn the $\mathcal{F}_{\text{systems}}$). 

We shall assume that there are $P \geq 1$ policies under which these traces
where observed. In particular, each trajectory was observed under one of these
$P$ policies and the assignment of policy to the trajectory was done uniformly
at random. Define $\Pi_p \subset [U]$ as collection of indices corresponding 
to trajectories $i \in [N]$ and their times $t \leq H_i$ where trajectory $i$
was assigned policy $p$ for $p \in [P]$. Let $U_p = |\Pi_p|$.

\medskip
\noindent{\bf Tensor factorization, low CP-rank.} The tensor $M$
admits (not necessarily unique) factorization of the form: for any $\alpha \in [A], 
\beta \in [U], \gamma \in [D]$
\begin{align}\label{apx.eq:factor}
    m_{\alpha \beta \gamma} & = \sum_{\ell=1}^r x_{\alpha \ell} y_{\beta \ell} z_{\gamma \ell},
\end{align}
for some $r \geq 1$. For any tensor, such a factorization exits with $r$ at most
${\sf poly}(A, U, D)$. 

\medskip\noindent{\em Assumption 1 (\textbf{low-rank factorization}).} We shall make an assumption that $r$ is {\em small}, i.e. 
does not scale with $A, U, D$ and specifically a small constant. 

\medskip\noindent{\em Assumption 2 (\textbf{sufficient measurements}).} We shall assume that number of
measurements per instance, $D$, is at least as large as the underlying rank $r$ of the tensor $M$, i.e. 
$D \geq r$. 

\medskip
\noindent{\bf Distributional invariance and RCT.} As before, we shall assume
that the distribution of latent factors is the same across different policies
due to random assignment of policies to trajectories in the setup of RCT. In
the context of the tensor $M$, this corresponds to the distribution invariance
of factors $y_{\beta \cdot} \in \mathbb{R}^r$ over $\beta \in \Pi_p$ for any $p \in [P]$.
Concretely, for any $p \neq p' \in [P]$ and $\ell \in [r]$, we have 
\begin{align}\label{apx.eq:inv}
    \frac{1}{U_p} \sum_{\beta \in \Pi_p} y_{\beta \ell} & \approx 
         \frac{1}{U_{p'}} \sum_{\beta' \in \Pi_{p'}} y_{\beta' \ell}. 
\end{align}
More generally, any finite moment (not just first moment or average) of latent factors should be empirically invariant across policies.  As in \S\ref{sec:MatrixCompletion}, we would like to utilize property \eqref{apx.eq:inv}
to estimate the tensor $M$.

\medskip
\noindent{\bf A Simple Estimation Method and When It Works.} We describe a simple method that can recover entire tensor as long as rank $r \leq D$. For simplicity, 
we shall assume $r = D$ (the largest possible rank for which method will work). 
By \eqref{apx.eq:factor}, for a given fixed $\alpha \in [A]$ and
across $\beta \in [U], \gamma \in [D]$, 
\begin{align}\label{apx.eq:factor.1}
    m_{\alpha \beta \gamma} & = 
    \sum_{\ell=1}^r  y_{\beta \ell} \tilde{z}^\alpha_{\gamma \ell},
\end{align}
where $\tilde{z}^\alpha_{\gamma \ell} = x_{\alpha \ell} z_{\gamma \ell}$. 
Since $D = r$, the matrix $\tilde{Z}^\alpha = [\tilde{z}^\alpha_{\gamma \ell}: 
\gamma \in [D], \ell \in [r]]$ is a square matrix. 
With this notation, we have that for any fixed $\alpha\in [A]$, the 
matrix $M^\alpha = [m_{\alpha \beta \gamma}: \beta \in [U], \gamma \in [D]] \in \mathbb{R}^{U \times D}$ (or
$\mathbb{R}^{U \times r}$ since $r = D$) can be represented as 
\begin{align}\label{eqn:matrix_factorization}
    M^\alpha & = Y \tilde{Z}^{\alpha, T}, 
\end{align}
where $Y = [y_{\beta \ell}: \beta \in [U], \ell \in [r]] \in \mathbb{R}^{U \times r}$.

\medskip
\noindent{\em Assumption 3 (\textbf{invertibility}).} We shall assume that 
the $D \times D$ (i.e. $r \times r$) square matrices $\tilde{Z}^\alpha$ 
for each $\alpha \in [A]$ are full rank and hence invertible. 

The {\em Assumption 3} implies that $Y = M^\alpha \big(\tilde{Z}^{\alpha, T}\big)^{-1}$
for all $\alpha \in [A]$. 

For policy $p \in [P]$, indices $\beta \in \Pi_p$ are relevant. For a given $\beta \in \Pi_p$, 
if the policy $p$ utilized action $\alpha \in [A]$, $m_{\alpha \beta \cdot} \in \mathbb{R}^D$ is
observed. To that end, let $\Pi_{p, \alpha} = \{\beta \in \Pi_p:$ policy utilized action $\alpha\}$. Let
$U_{p,\alpha} = |\Pi_{p, \alpha}|$ for any $\alpha \in [A]$.  Then, define 
$Y^{p, \alpha} = [y_{\beta \ell}: \beta \in \Pi_{p, \alpha}, \ell \in [r]] \in \mathbb{R}^{U_{p, \alpha} \times r}$, 
$M^{\alpha, p} = [m_{\alpha \beta \gamma}: \beta \in \Pi_{p, \alpha}, \gamma \in [D]]$. Then we
have $Y^{p, \alpha} = M^{\alpha, p} \big(\tilde{Z}^{\alpha, T}\big)^{-1}$. 

Therefore, for any $\ell \in [r=D]$, 
\begin{align}
\sum_{\beta \in \Pi_{p, \alpha}} y_{\beta \ell} & = \bOne^{p,\alpha, T} Y^{p,\alpha} \be_\ell \nonumber \\
 & = \be_\ell^T Y^{p,\alpha ,T}  \bOne^{p,\alpha}  \nonumber \\
& =   \be_\ell^T \big(\tilde{Z}^{\alpha}\big)^{-1} M^{\alpha, p, T}    \bOne^{p,\alpha},
\end{align}
where $\bOne^{p,\alpha} \in \mathbb{R}^{U_{p, \alpha}}$ is vector of all $1$s, and $\be_\ell \in \mathbb{R}^{r}$ be
vector with all entries $0$ but the $\ell\in [r]$th co-ordinate $1$ . 

Then, for any $\ell \in [r]$ and $p \in [P]$,
\begin{align}\label{apx.eq:inv.1}
\frac{1}{U_p} \sum_{\beta \in \Pi_{p}} y_{\beta \ell} & = \frac{1}{U_p}  \sum_{\alpha \in [A]}  \sum_{\beta \in \Pi_{p, \alpha}} y_{\beta \ell} \nonumber \\
& = \frac{1}{U_p}  \sum_{\alpha \in [A]}   \be_\ell^T \big(\tilde{Z}^{\alpha}\big)^{-1} M^{\alpha, p, T}    \bOne^{p,\alpha}
\nonumber\\
& =  \sum_{\alpha \in [A]}   \be_\ell^T \big(\tilde{Z}^{\alpha}\big)^{-1} \Big( \frac{1}{U_p}  M^{\alpha, p, T}    \bOne^{p,\alpha} \Big)
\nonumber\\
& =  \sum_{\alpha \in [A]}   \be_\ell^T \big(\tilde{Z}^{\alpha}\big)^{-1} {\mathcal M}^{\alpha, p} , 
\end{align} 
where ${\mathcal M}^{\alpha, p} =  \frac{1}{U_p}  M^{\alpha, p, T}    \bOne^{p,\alpha}   \in \mathbb{R}^{r,1}$ is
an observed quantity, while $\tilde{Z}^{\alpha, T}$ is unknown. Using \eqref{apx.eq:inv.1} and \eqref{apx.eq:inv}, we obtain that for any $\ell \in [r]$
and $p \neq p' \in [P]$,
\begin{align}\label{apx.eq:inv.2}
\sum_{\alpha \in [A]}   \be_\ell^T \big(\tilde{Z}^{\alpha}\big)^{-1} {\mathcal M}^{\alpha, p} & \approx \sum_{\alpha \in [A]}   \be_\ell^T \big(\tilde{Z}^{\alpha}\big)^{-1} {\mathcal M}^{\alpha, p'}.
\end{align} 
Let $\tilde{z}^{\alpha, \ell} =  \be_\ell^T \big(\tilde{Z}^{\alpha}\big)^{-1}  \in \mathbb{R}^{1,r}$ be the $\ell$th row the of $r\times r$ 
matrix $\big(\tilde{Z}^{\alpha}\big)^{-1}$. Then \eqref{apx.eq:inv.2} implies that for any $\ell \in [r]$
and $p \neq p' \in [P]$,
\begin{align}\label{apx.eq:inv.3}
    \sum_{\alpha \in [A]} \tilde{z}^{\alpha, \ell}  ({\mathcal M}^{\alpha, p} - {\mathcal M}^{\alpha, p'}) & \approx 0. 
\end{align}

Which can be written in matrix form as 

\begin{align}
\begin{bmatrix}
 \tilde{z}^{1, \ell}  &
 \tilde{z}^{2, \ell} &
\dots &
 \tilde{z}^{A, \ell} 
\end{bmatrix}
\begin{bmatrix}
{\mathcal M}^{1, p} - {\mathcal M}^{1, p'}\\
{\mathcal M}^{2, p} - {\mathcal M}^{2, p'} \\
\vdots \\
{\mathcal M}^{A, p} - {\mathcal M}^{A, p'}\\
\end{bmatrix} = 0
\end{align}
By noting that that this hold for all $\ell \in [r]$, and recalling that  $\tilde{z}^{\alpha, \ell}$  is the $\ell$-th row the of the $r\times r$ 
matrix $\big(\tilde{Z}^{\alpha}\big)^{-1}$, we get, 
\begin{align} \label{apx.eq:inv.4}
\begin{bmatrix}
 \big(\tilde{Z}^{1}\big)^{-1} &
\big(\tilde{Z}^{2}\big)^{-1}&
\dots &
\big(\tilde{Z}^{A}\big)^{-1}
\end{bmatrix}
\begin{bmatrix}
{\mathcal M}^{1, p} - {\mathcal M}^{1, p'}\\
{\mathcal M}^{2, p} - {\mathcal M}^{2, p'} \\
\vdots \\
{\mathcal M}^{A, p} - {\mathcal M}^{A, p'}\\
\end{bmatrix} = \bf{0}, 
\end{align}
where $\bf{0}$ is a vector of zeros of size $r$. Note that the above is a system of $r$ linear equations, with $Ar^2$ unknowns (recall that the  $r\times r$ 
matrices $\big(\tilde{Z}^{\alpha}\big)^{-1}$ are unknown for $\alpha \in[A]$ ). Let $ {\bf Z}   \in \mathbb{R}^{r \times Ar} $ and  ${\bf v}^{p, p'}  \in \mathbb{R}^{Ar}$ denote the first and second matrix in the left hand side, respectively, 
then  \eqref{apx.eq:inv.4} can be re-written as, 
\begin{align}\label{apx.eq:inv.5}
 {\bf Z} {\bf v}^{p, p'} & \approx \mathbf{0}.
\end{align} 
By definition,  ${\bf v}^{p, p'}$ is observed quantity for each $p \neq p' \in [P]$. Now
if we consider $P-1$ equations produced by considering pair of policies $(1,2), (1,3), \dots, (1, P)$ in \eqref{apx.eq:inv.5}, by design they 
are non-redundant linear equations. Let matrix ${\bf V} \in \mathbb{R}^{Ar \times P-1}$ be formed by stacking 
${\bf v}^{1,2}, \dots, {\bf v}^{1,P}$ column-wise. 

Furthermore, let us define $\mathbf{s}^p \in \mathbb{R}^{Ar}$ as $[\mathcal{M}^{1,p}, \cdots, \mathcal{M}^{A,p}]^\intercal$. 
Define $\mathbf{S} \in \mathbb{R}^{Ar \times P}$ by stacking $\mathbf{s}^1, \cdots, \mathbf{s}^P$ column-wise.

\medskip
\noindent{\em Assumption 4 (\textbf{Sufficient, Diverse Policies}).} Let $P \geq Ar$ and the rank of ${\bf S} = Ar$. 

Note that we can derive $\mathbf{V}$ from $\mathbf{S}$ by subtracting the first column from all other columns, and removing the first column. Thus, Under Assumption 4, the +rank of $\mathbf{V}$ is at least $Ar-1$. Further, given Assumption 3 which excludes the scenario ${\bf Z} =0$, it follows that the rank of $\mathbf{V}$ is $Ar-1$.
As rank of $\mathbf{V}$ is $Ar-1$, we can uniquely (up to scaling) recover ${\bf Z}$ by solving for system of linear equation 
${\bf Z}  {\bf V} = {\bf 0}$ as the null space of ${\bf V}$ is of dimension $1$.

Once we know ${\bf z}$, i.e. by undoing flattening, we obtain $\big(\tilde{Z}^{\alpha, T}\big)^{-1}$ for each $\alpha \in [A]$. 
Since for each policy $p \in [P]$ and $\alpha \in [A]$, $Y^{p, \alpha} = M^{\alpha, p} \big(\tilde{Z}^{\alpha, T}\big)^{-1}$ 
and we observe $M^{\alpha, p}$, we can recover $Y^{p, \alpha}$ and hence subsequently $Y \in \mathbb{R}^{U \times r}$. 

By \eqref{eqn:matrix_factorization}, we can now recover slice of tensor $M$, the $M^\alpha$ for each $\alpha \in [A]$, and hence 
we can recover entire tensor $M$ as desired. 

\noindent \textbf{Interpretation of Assumption 4.}
Consider $\beta^{\text{th}}$ Column of the matrix $\mathbf{S}$, i.e., 
$\big[
\mathbb{E}[m^\intercal|i=1,\pi_{\beta}]\mathbb{P}(i=1|\pi_{\beta}),
\cdots,
\mathbb{E}[m^\intercal|i=A,\pi_{\beta}]\mathbb{P}(i=A|\pi_{\beta})
\big]^\intercal$
where $i$ denotes the action index and $\beta$ the policy index.
This column is a vector of statistics associated with traces collected using policy $\beta$.
Each element in this vector consists of two components: the first component is the conditional mean of the trace given a specific action, and the second element is the probability of taking this action.
We interpret linear independence of each of these components for different policy vectors as policy diversity.
For instance, think of the second component which captures probability vectors of different actions for each policy. 
Its linear independence across different policies roughly means that each policy should assign new probability vectors to different actions, and not a probability vector similar (linearly dependent) to that of previous policies.
Also note that this assumption is not satisfied if an action is not taken by any of the policies which makes all elements of the corresponding row equal to zero.
\clearpage
\section{Real-world ABR}
\label{app:puffer}

\subsection{Comprehensive results}
\label{app:puffer:all}

\input{sections/figures/app_puffer_all_cdf}

In \Cref{fig:puffer:emd_cdf}, we presented a concise view of simulator fidelity, for an internal variable in \gls{abr} sessions called buffer occupancy level. Specifically, we considered the simulation of a target policy, given trajectories collected using a different source policy. We measured the error between buffer simulations and ground truth through \gls{emd}, a similarity index for distributions. For a complementary view, we provide the full distributions in \Cref{app:fig:puffer:all}, for all simulators and ground truth for target and source policies. Below each plot, we also report the \gls{emd} of \CausalSim predictions.

\subsection{Policy Discriminator and \\ Latent Invariance}
\label{app:puffer_confusion}

The policy discriminator ($\mathcal{W}_{\gamma}$ in \Cref{fig:overview_comb}) described in \S\ref{sec:solution} has the goal of predicting the source policy, given a latent factor generated by the latent factor extractor ($\mathcal{E}_{\theta}$ in \Cref{fig:overview_comb}). Since our data is collected with an RCT, the true latent factor distribution should be indifferent to the source policy. Therefore, if the latent factor extractor generates the ground truth latent factors, the policy discriminator should not be able to predict the source policy accurately. In fact, even the optimal policy discriminator outputs the population share of each source policy (e.g. what fraction of the data comes from BBA) in the training data~\cite{goodfellow2014generative}. To assess this statement, we present the confusion matrix and population share of source data, for three left-out policies in \Cref{table:puffer_conf_mats}. Each row corresponds to one source policy, and each column corresponds to the policy discriminator's prediction of the source policy. We observe that predictions do not change noticeably with different source policies, and that they closely match the population share for each left-out policy. This demonstrates that the extracted latent features were indeed invariant to the source policy.

\begin{table}
\begin{subtable}{\linewidth}
    \small
    \centering
    \begin{tabular}{lcccc}
    \toprule
    & \multicolumn{4}{c}{\textbf{Prediction}}\\
    \cmidrule{2-5}
	\textbf{Source Policy} & BOLA2 & BOLA1 & Fugu-CL & Fugu-2019 \\
	\midrule
	BOLA2 & 22.44\% & 22.58\% & 26.99\% & 27.99\%\\
	BOLA1 & 22.43\% & 22.58\% & 26.99\% & 27.99\%\\
	Fugu-CL & 22.44\% & 22.58\% & 26.99\% & 27.99\%\\
	Fugu-2019 & 22.44\% & 22.58\% & 26.99\% & 28.00\%\\
	\midrule
	& \multicolumn{4}{c}{\textbf{Source Policy}}\\
	\cmidrule{2-5}
	& BOLA2 & BOLA1 & Fugu-CL & Fugu-2019\\
	\cmidrule{2-5}
	\textbf{Population} & 22.45\% & 22.50\% & 27.11\% & 27.94\%\\
    \bottomrule
    \end{tabular}
    \caption{Left-out policy is BBA}
    \label{subtab:conf_mat_bba}
    \vspace{0.5cm}
\end{subtable}
\begin{subtable}{\linewidth}
    \small
    \centering
    \begin{tabular}{lcccc}
    \toprule
    & \multicolumn{4}{c}{\textbf{Predictions}}\\
    \cmidrule{2-5}
	\textbf{Source Policy} & BOLA2 & Fugu-CL & Fugu-2019 & BBA \\
	\midrule
	BOLA2 & 21.34\% & 26.04\% & 26.75\% & 25.87\%\\
	Fugu-CL & 21.33\% & 26.05\% & 26.75\% & 25.87\%\\
	Fugu-2019 & 21.33\% & 26.04\% & 26.77\% & 25.86\%\\
	BBA & 21.33\% & 26.04\% & 26.76\% & 25.87\%\\
	\midrule
	& \multicolumn{4}{c}{\textbf{Source Policy}}\\
	\cmidrule{2-5}
	& BOLA2 & Fugu-CL & Fugu-2019 & BBA\\
	\cmidrule{2-5}
	\textbf{Population} & 21.48\% & 25.94\% & 26.74\% & 25.84\%\\
    \bottomrule
    \end{tabular}
    \caption{Left-out policy is BOLA1}
    \label{subtab:conf_mat_bola1}
    \vspace{0.5cm}
\end{subtable}
\begin{subtable}{\linewidth}
    \small
    \centering
    \begin{tabular}{lcccc}
    \toprule
    & \multicolumn{4}{c}{\textbf{Predictions}}\\
    \cmidrule{2-5}
	\textbf{Source Policy} & BOLA1 & Fugu-CL & Fugu-2019 & BBA \\
	\midrule
	BOLA1 & 21.46\% & 26.00\% & 26.76\% & 25.78\%\\
	Fugu-CL & 21.45\% & 26.01\% & 26.77\% & 25.76\%\\
	Fugu-2019 & 21.45\% & 26.00\% & 26.79\% & 25.76\%\\
	BBA & 21.45\% & 25.99\% & 26.76\% & 25.80\%\\
	\midrule
	& \multicolumn{4}{c}{\textbf{Source Policy}}\\
	\cmidrule{2-5}
	& BOLA1 & Fugu-CL & Fugu-2019 & BBA\\
	\cmidrule{2-5}
	\textbf{Population} & 21.52\% & 25.93\% & 26.72\% & 25.83\%\\
    \bottomrule
    \end{tabular}
    \caption{Left-out policy is BOLA2}
    \label{subtab:conf_mat_bola2}
\end{subtable}
\caption{Confusion matrix and population statistics for the policy discriminator with three left out policies.}
\label{table:puffer_conf_mats}
\end{table}

\subsection{What makes a simulation \\ scenario easy/hard?}
\label{app:sub:difficulty}
In \S\ref{subsec:buffer_eval}, we compared the accuracy of \CausalSim, ExpertSim and SLSim, in a simulation task on real \gls{abr} data. We observed that in about 30\% of scenarios, which we call \emph{easy} scenarios, all simulators perform well. However, in about 70\% of the source/target scenarios, which we call \emph{hard} simulation scenarios, baseline predictions are highly biased towards the source distributions. In these hard scenarios, \CausalSim is able to de-bias the trajectories and its predictions match the target distribution well, as observable in \Cref{app:fig:puffer:all}.

So it is natural to wonder what makes a simulation scenario easy/hard? An easy simulation scenario happens when source and target policies take similar actions. 
Similar action means that the factual achieved throughput (of the source policy) is similar to the counterfactual achieved throughput (of the target policy).
This is what both ExpertSim (explicitly) and SLSim (implicitly) assume for doing simulation. 
Making this assumption is the core reason their simulations are biased in hard cases, where source and target policies take different actions, as we discussed in detail in \S\ref{ssec:minRTT}.

\begin{figure}[!bt]
    \centering
    \includegraphics[width=\linewidth]{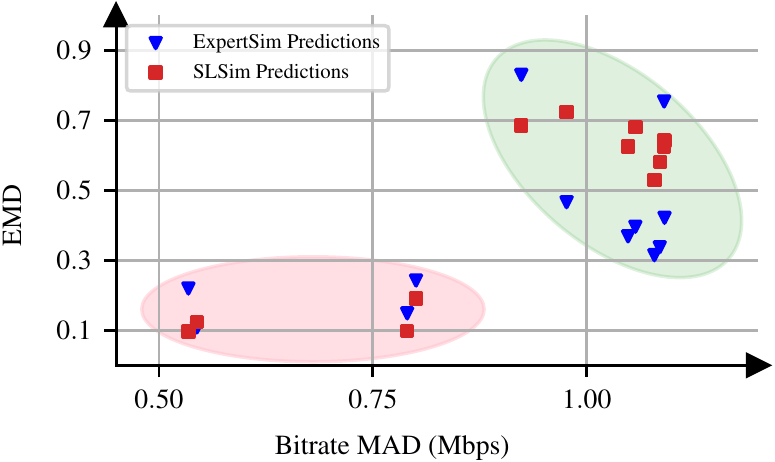}
    \caption{Simulation difficulty is related to how different counterfactual actions are from factual ones. This figure shows scatterplot of EMD versus mean absolute bitrate difference, for ExpertSim and SLSim, over all possible source left-out pairs. The pink cluster signifies the `easy' scenarios and the green cluster signifies `hard' ones.}
    \label{app:fig:puffer_difficulty}
\end{figure}

\Cref{app:fig:puffer_difficulty} validates our reasoning for what makes a simulation scenario difficult.
The X axis shows the Mean Absolute Difference (MAD) between source and simulation actions (bitrates) when simulating with SLSim in a specific source/target scenario.
Y axis shows \gls{emd} (Our performance metric for simulation, smaller is better) of both baselines in that specific scenario.

Two main cluster of points are clearly visible in this figure.
The pink cluster on the bottom left corresponds to easy simulations.
It includes all source/target simulation scenarios where baselines perform well (bottom), and at the same time, source and target actions are quite similar (left).

The green cluster at the top right corresponds to the hard simulations.
It includes all source/target simulation scenarios where baselines fail to perform an unbiased simulation (top), and at the same time, source and target actions are quite different (right).

\subsection{A More Fine-grained \\  Evaluation}\label{sec:fine_grained}

Ideally, we would like to evaluate \CausalSim's simulation to ground truth on a step-by-step basis for a given trajectory.
But as discussed in \S\ref{subsec:buffer_eval}, this is not possible in real-world data, as we only see the outcome of one \gls{abr} algorithm's chosen action for a single step.
In other words, there is no way to get ground truth for individual steps in the observational data, which is referred to as the fundamental problem of Causal Inference~\cite{holland}. 
This is the reason we evaluated predictions on a distributional level.

However, there is a way to evaluate \CausalSim's predictions at a more fine-grained level. 
Instead of evaluating the predicted distribution of buffer occupancy across the whole population, we can evaluate on certain {\em sub-populations} of users. The only requirement is that the way we select these sub-populations should be statistically independent of the \gls{abr} algorithm.
For example, we can partition users by a metric such as Min \gls{rtt}, which is independent of the policy chosen for each user in the RCT. Min \gls{rtt} is an inherent property of a network path\footnote{This is true to a first order approximation, if we ignore the possibility that a video streaming session drives up queueing delays throughout the course of a video, thereby inflating the observed Min RTT.}, and we would expect Min RTT distribution to be the same for users assigned to different ABR policies. 

We use the Min\gls{rtt} to create the following four sub-populations:
\begin{enumerate}
    \item Sub1: users with $\text{Min \gls{rtt}} < 35^{ms}$
    \item Sub2: users with $35^{ms} \le \text{Min \gls{rtt}} < 70^{ms}$
    \item Sub3: users with $70^{ms} \le \text{Min \gls{rtt}} < 100^{ms}$
    \item Sub4: users with $100^{ms} \le \text{Min \gls{rtt}}$
\end{enumerate}
Now, we can ask question of the following type: \emph{had the users in sub-population two, who were assigned the source \gls{abr} algorithm, instead used the left-out \gls{abr} algorithm, what would the distribution of their buffer level look like?}
As the ground truth answer to this question, we can use the buffer level distribution of users in sub-population two assigned to the left-out policy.

\Cref{fig:puffer:cdf_subpop} shows the CDF of \CausalSim's \gls{emd} when simulating the left-out \gls{abr} algorithm over each of the above sub-populations.
We can see that \CausalSim maintains a superior \gls{emd} CDF compared to ExpertSim and SLSim, and remains accurate across different sub-populations. This further suggests that even at surgically small subpopulations, \CausalSim maintains accuracy, and does not overfit to the whole distribution.

\subsection{How to Tune \CausalSim's \\ Hyper-parameters?}
\label{sec:puffer_hyper_params}

\begin{figure}
    \begin{subfigure}{0.45\textwidth}
        \centering
        \includegraphics[width=\linewidth]{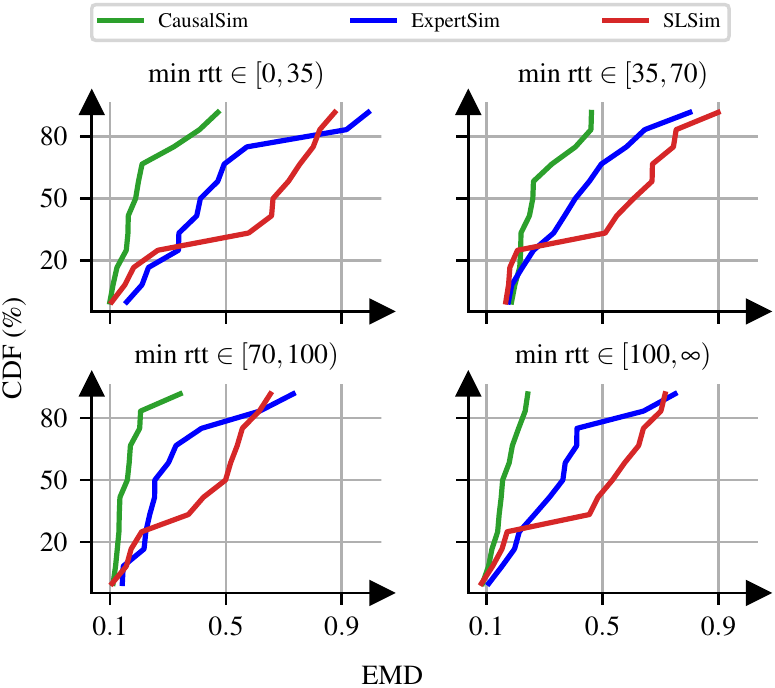}
        \caption{}
        \label{fig:puffer:cdf_subpop}
    \end{subfigure}
    \begin{subfigure}{0.45\textwidth}
        \centering
        \hspace*{-0.03\linewidth}
        \begin{tikzpicture}
            \begin{axis}[height=5cm,
                         width=\linewidth,
                         axis lines=left,
                         xlabel={Validation \gls{emd}},
                         ylabel={Test \gls{emd}},
                         font=\small,
                         enlargelimits=true,
                         xmax=2.5,
                         xmin=0,
                         ymax=3, 
                         ymin=0,
                         ylabel near ticks,
                         xlabel near ticks,
                         grid=both
                         ]
                \input{figures/raw_data/puffer/new_test_train_corr}
            \end{axis}
        \end{tikzpicture}
        \caption{}
        \label{fig:puffer:corr}
    \end{subfigure}
    \caption{\textbf{(a)} Comparing the distribution of \CausalSim \gls{emd}s with ExpertSim and SLSim over different sub-populations. \textbf{(b)} Validation \gls{emd} and test \gls{emd} are highly correlated. This justifies our hyper-parameter tuning strategy.}
\end{figure}

Counterfactual prediction is not a standard supervised learning task that optimizes in-distribution generalization.
Rather, it is always an \gls{ood} generalization problem, i.e., we collect data from a training policy (distribution 1), and want to accurately simulate data under a different policy (distribution 2).
Since we do not use data from the test policy when we train \CausalSim, we use the following natural proxy for tuning hyper-parameters:
\emph{Simulating \gls{abr} algorithms in the training data using trajectories of other \gls{abr} algorithms in the training data.}
This of course can be viewed as an \gls{ood} problem as well.
We claim that if a choice of hyper-parameters results in a robust model that performs well \gls{ood} across all validation \gls{abr} algorithms in the training data, it should work well for the actual left-out test policy as well.

We verify this hyper-parameter tuning procedure empirically.
For each choice of the three left-out \gls{abr} algorithms (hence training dataset), we train eleven different \CausalSim models with different choices of $\kappa$ (defined in \Cref{eqn:total_loss}).
We consider two metrics: \emph{(i) Test \gls{emd}}, defined as the average \gls{emd} when simulating the left-out \gls{abr} algorithm with trajectories in the training dataset. This is our main performance objective. 
\emph{(ii) Validation \gls{emd}}, defined as the average \gls{emd} when simulating \gls{abr} algorithms in the training dataset with trajectories in the training data that were collected with other \gls{abr} algorithms. This is our proxy objective for hyper-parameter tuning.

For each model (33 in all: 3 datasets, 11 example hyper-parameters), we calculate both Test \gls{emd} and Validation \gls{emd}, which results in one (Validation \gls{emd}, Test \gls{emd}) point in \Cref{fig:puffer:corr}.
The \gls{pcc} between Valid \gls{emd} and Test \gls{emd} is $0.92$, which shows high linear correlation.
Hence, though \CausalSim might not always perform well (i.e., Test \gls{emd} is not low for some combinations of training dataset and hyper-parameters), we can have a very good idea of how well it works by measuring Validation \gls{emd}.

\subsection{How to Tune SLSim's \\ Hyper-parameters?}
\label{sec:slsim_hyper_params}
SLSim takes as input the current buffer value, selected chunk size and observed throughput, and similar to \CausalSim, predicts the next buffer $\hat{b}_{t+1}$ and download time $\hat{d}_t$. We add two knobs to tune while training SLSim: \textbf{(1)} The loss function $\mathcal{L}_{\xi}(\cdot, \cdot)$ used to steer the \gls{nn} output to the ground truth output, and \textbf{(2)} The relative weighting of the loss function for download time with respect to that of the buffer occupancy, $\eta$. Concretely, we use the following total loss:

\begin{equation}
\label{eqn:slsim_total_loss}
    {\mathcal{L}}_{\text{slsim}} = \mathbb{E}_B \left[ \frac{1}{\eta + 1}.\mathcal{L}_{\xi}(\hat{b}_{t+1}, b_{t+1}) + \frac{\eta}{\eta + 1}.\mathcal{L}_{\xi}(\hat{d}_{t}, d_{t}) \right]
\end{equation}

where the expectation is over the a sampled minibatch $B$ from dataset $D$, and $b_{t+1}$ and $d_{t}$ denote the ground truth values for next buffer level and chunk download time. \Cref{table:puffer_train} lists the loss functions and $\eta$ values considered.

To tune these values, we use ground truth data from all policies except a left out policy. We then proceed with the proxy tuning objective used in \S\ref{sec:puffer_hyper_params}, i.e. we look for the configuration with the highest accuracy at simulating algorithms in the training data using trajectories of other algorithms in the training data. We then use the resulting configuration (and model) to simulate the left-out policy on the training data.

From the perspective of tuning, this methodology puts SLSim on equal ground with respect to \CausalSim, and makes for a fair comparison. Note that we do not tune loss function type or $\eta$ with \CausalSim due to limited computational resources, but tuning those as well could potentially improve \CausalSim's accuracy.

\subsection{Simulation Accuracy: Continued}
\label{app:subsec:puffer_separated}
\begin{figure}
    \centering
    \begin{subfigure}{\linewidth}
        \centering
        \includegraphics[width=\linewidth]{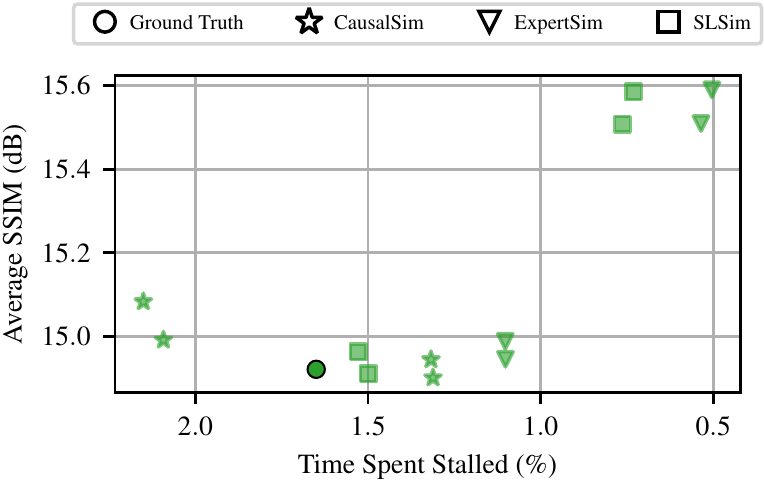}
        \caption{}
        \label{fig:bba_partial_scatter}
    \end{subfigure}
    \begin{subfigure}{\linewidth}
        \centering
        \includegraphics[width=\linewidth]{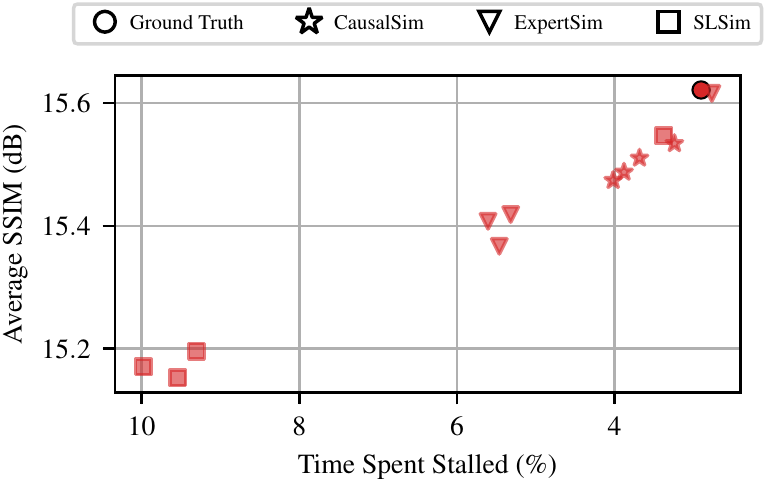}
        \caption{}
        \label{fig:bola2_partial_scatter}
  \end{subfigure}
    \caption{Predictions for \textbf{(a)} BBA and \textbf{(b)} BOLA2, separated by the ABR algorithm source data was collected with. Each point indicates a specific \textbf{source} ABR algorithm.}
\end{figure}

In \S\ref{subsubsec:results:puffer:can_sim}, we stated that ExpertSim and SLSim predictions are significantly affected by the source data they are simulating on, and demonstrated the effect of source policies on BOLA1 predictions in \Cref{fig:bola1_partial_scatter}. Here, we demonstrate the same figure for BBA in \Cref{fig:bba_partial_scatter} and BOLA2 in \Cref{fig:bola2_partial_scatter}. \CausalSim is designed to remove the bias of the algorithm used for collecting source data when simulating a target policy and its predictions remains unaffected by the performance of that source policy. ExpertSim and SLSim however, due to the violation of the exogenous trace assumption, will predict different metrics when using different source traces.

\subsection{Dataset \& Algorithms}
\label{app:puffer:alg_data}

\begin{table*}
\small
\centering
\begin{tabular}{ c l l c c }
\toprule
\textbf{Policies} & \textbf{Hyperparameter} & \textbf{Value} & Used as source & Used as left out \\
\midrule
\multirow{2}{*}{\textbf{\gls{bba}}} & Cushion & 3 (as used in puffer) & \multirow{2}{*}{\checkmark} &  \multirow{2}{*}{\checkmark} \\
\cmidrule{2-3}
& Reservoir & 10.5 (as used in puffer) & &\\
\midrule
\multirow{5}{*}{\textbf{BOLA-BASIC v1}} & $V$ & 0.67 (As computed in puffer) & \multirow{5}{*}{\checkmark} &  \multirow{5}{*}{\checkmark} \\
\cmidrule{2-3}
& $\gamma$ & -0.43 (As computed in puffer) & &\\
\cmidrule{2-3}
& Utility function & $\log_{10}(1-ssim)$ (As used in puffer) & &\\
\cmidrule{2-3}
& Minimum utility & 0 dB (As used in puffer) & &\\
\cmidrule{2-3}
& Maximum utility & 60 dB (As used in puffer) & &\\
\midrule
\multirow{5}{*}{\textbf{BOLA-BASIC v2}} & $V$ & 51.4 (As computed in puffer) & \multirow{5}{*}{\checkmark} &  \multirow{5}{*}{\checkmark} \\
\cmidrule{2-3}
& $\gamma$ & -0.43 (As computed in puffer) & &\\
\cmidrule{2-3}
& Utility function & $ssim$ (As used in puffer) & &\\
\cmidrule{2-3}
& Minimum utility & 0 (As used in puffer) & &\\
\cmidrule{2-3}
& Maximum utility & 1 (As used in puffer) & &\\
\midrule
\textbf{Fugu-CL} & - & - & \checkmark & $\times$ \\
\midrule
\textbf{Fugu-2019} & - & - & \checkmark &  $\times$ \\
\midrule
\bottomrule
\end{tabular}
\caption{\gls{abr} algorithms used in the real-world dataset and experiments}
\vspace{20pt}
\label{table:puffer_pol}
\end{table*}

Our trajectories in the real-world (Puffer) data come from `slow streams` in the time span of July 27, 2020 until June 2, 2021. In this period of time, 5 ABR algorithms appear consistently and are listed in \Cref{table:puffer_pol}. 
Each trajectory is an active client session streaming a live TV channel. 
We follow Puffer's definition of `slow streams'; streams with TCP delivery rates below 6 Mbps. We use `slow streams` data, since the highest quality chunks rarely surpass $6-7$ Mbps, and paths with higher bandwidth will always stream the highest quality chunks under all policies.
Puffer uses the same reasoning and evaluates algorithms at two population levels; 'slow streams' and 'all streams'.

In aggregating `slow stream` logs, we met several difficulties that we outline here for reproducibility. Data without these difficulties would potentially improve \CausalSim's accuracy. Note that this does not affect \Cref{fig:puffer_bayes:rct}, as the data for that figure is reported directly on Puffer~\cite{puffer_rct1, puffer_rct2}.

Puffer logs are reported as three separate event groups; 1) `video\_sent': the first packet of a chunk is sent, 2) `video\_acked': The last packet of a chunk is acknowledged, 3) `client': The client sent a message. Stall rate is computed using the `client' logs and quality is computed using the `video\_sent' logs. 
\begin{enumerate}
    \item To compute download time, we have to merge `video\_sent' and `video\_acked', and ensure that merged logs are consecutive in timestamps, i.e. no chunk is missing in between two other chunks. However, in the current data this removes all chunks that have been sent but not acknowledged, usually the last chunk. Puffer uses these chunks in measuring quality level, but we can't. This did not have any measurable impact, however.
    \item To compute stall rate, both total stall time and total watch time are computed with `client' logs. For this, the latest report that obeys a set of rules is used. We, however, have to compute stall time and watch time using our merged logs (merged logs are also what we get out of simulation). This would be easy on the original data, if `client` logs and `video\_sent' were in sync, but they are not; whenever a rebuffering is reported by the client, `client' log is updated but `video\_sent' is updated in the next few chunks. To circumvent this, we recompute rebuffering as $t_r=max(0, t_d-b)$, where $t_r$ is rebuffering, $b$ is buffer occupancy and $t_d$ is  download time.
    This formula is off by half of an RTT, and empirically inflates stall rates by $1.26-1.31x$, for all policies. In the absence of synchronized data, this is the best we can recover, but it does not affect the comparison among policies. Hence, we believe simulating with this data should lead to similar trends as with clean unperturbed data.
    \item We cannot calculate watch time as Puffer does, since we have to use the merged log. We tried several simple formulas that should calculate watch time, but oddly most turn out to be inaccurate. One reason is that in some streams, buffer playback rate is not 1, i.e. one second of buffer is not depleted per second. These streams are likely due to browser tabs put in background, and throttled by the browser threading system. As a workaround, we use the original watch time minus the original stall time that Puffer computed for a stream, and offset it by the total stall time in the simulation.
    \item At each step, the buffer should not increase by more than a single chunk, $2.002$ seconds, but it does (sometimes by as much as 14 seconds). We filter such data out.
    \item When we are about to send a chunk, our last reported buffer value must never dip below 2.002 (except in the beginning). When buffer is below 15 seconds, the next chunk must be sent immediately after the last one. If rebuffering occurs, the next buffer value will be exactly $2.002$ and if it doesn't, it will be larger than $2.002$. We frequently (more than one million instances) observe buffer values below $2.002$. We do not filter them out, as this would invalidate most logs. 
\end{enumerate}

To test out \CausalSim, we need to simulate the streaming session using a different algorithm than the one that was actually used in that session. This requires implementation of the ABR algorithms. 
To ensure our implementations are correct, we attempt to reconstruct the choices made at runtime by each policy, and compare them to the logged choices. We expect our reproduction to match 100\% when our implementation is faithful and logs match runtime inputs. For the logs in July 27th, 2020, we observe 100\% matching for BOLA1 and BOLA2 and 99.993\% for BBA. For the latter, there are rare cases where two encodings are seemingly equal in SSIM up to the 6 logged decimal places, but were likely slightly different in double precision format at runtime. These instances are rare enough that we can ignore them.

For Fugu-2019 or Fugu-CL however, our reproductions did not match in 6\% and 19\% of cases, whether we used the original C implementation or our own Python port. 
The Puffer team informed us of a use-after-free issue regarding the \gls{tcp} info struct that was fixed in March 7th, 2022. Hence we retried this process for the logs pertaining to July 27th, 2022 and the error rate shrank to 0.53\% and 0.64\%. Unfortunately, a 0.5\% error rate is still too high and even if we ignore that, limits us to RCT logs after March 7th.
Therefore, we do not consider Fugu-2019 or Fugu-CL as candidates for left-out algorithms.

\subsection{Training setup}
\label{app:puffer:train}

\begin{table*}
\small
\centering
\begin{tabular}{ c l l }
\toprule
\textbf{Model} & \textbf{Hyperparameter} & \textbf{Value} \\
\midrule
\multirow{12}{*}{\textbf{SLSim} (1 network), \textbf{\CausalSim} (3 networks)} & Hidden layers & (128, 128) \\
\cmidrule{2-3}
& Hidden layer Activation function & \gls{relu} \\
\cmidrule{2-3}
& Output layer Activation function & Identity mapping \\
\cmidrule{2-3}
& Optimizer & Adam \cite{adam} \\
\cmidrule{2-3}
& Learning rate & 0.001 \\
\cmidrule{2-3}
& $\beta_1$ & 0.9 \\
\cmidrule{2-3}
& $\beta_2$ & 0.999 \\
\cmidrule{2-3}
& $\epsilon$ & $10^{-8}$ \\
\cmidrule{2-3}
& Batch size & $2^{17}$ \\
\midrule
\multirow{8}{*}{\textbf{\CausalSim}} & \multirow{2}{*}{$\kappa$} & \{0.05, 0.1, 0.5, 1, 5,  \\
& & 10, 15, 20 ,25, 30, 40\} \\
\cmidrule{2-3}
& Training iterations (num\_train\_it) & 5000 \\
\cmidrule{2-3}
& num\_disc\_it & 10 \\
\cmidrule{2-3}
& Loss function & Huber($\delta=0.2$) \\
\cmidrule{2-3}
& $\eta$ (download time weight wrt buffer) & 1 \\
\midrule
\multirow{4}{*}{\textbf{SLSim}} & Training iterations & 10000 \\
\cmidrule{2-3}
& Loss function & \{Huber($\delta=0.2$), L1, MSE\} \\
\cmidrule{2-3}
& $\eta$ (download time weight wrt buffer) & \{0.5, 1, 10\} \\
\midrule
\bottomrule
\end{tabular}
\caption{Training setup and hyperparameters for the real-world \gls{abr} experiment}
\label{table:puffer_train}
\end{table*}

We use \glspl{mlp} as the \gls{nn} structures for \CausalSim models and the SLSim model. All implementations use the Pytorch~\cite{pytorch} library. \Cref{table:puffer_train} is a comprehensive list of all hyperparameters used in training. 

\section{Synthetic ABR}
\label{sec:abr_synth}

As explained in \S\ref{subsub:puffer_additional_exp}, we also evaluate \CausalSim in a synthetic ABR environment, in which we can obtain ground truth for individual counterfactual predictions on a step-by-step basis for a trajectory. In these experiments, we also use a larger set of policies than available in the real data.

\subsection{Simulation Dynamics}
\label{app:abr_synth:dynamics}
In each simulated training session, we start with an empty playback buffer and a latent network path characterized by an \gls{rtt} and a capacity trace. 
In each step, an \gls{abr} algorithm chooses a chunk size, which is transported over this network path to the client as the buffer is depleting. 
Once the user receives the chunk, the buffer level increases by the chunk duration. 
This simple system can be modeled as follows:
\begin{equation}
    b_{t+1} = min(b_t - d_t, 0) + c
\end{equation}
where $b_t$, $d_t$ and $c$ refer to the buffer level at time step $t$, the download time of the chunk at time step $t$, and the chunk video length in seconds, respectively. 
Streaming the next chunk is started immediately following receiving the previous one, except when the buffer level surpasses a certain value (in our case, 10 seconds to mimic a live-stream \gls{abr} setting). 
To compute $d_t$, we model the transport as a \gls{tcp} session with an \gls{aimd} congestion control mechanism with slow start. 
For every chunk, the \gls{tcp} connection starts from the minimum window size of 2 packets and increases the window according to slow start. 
Therefore, it takes the transport some time to begin fully utilizing the available network capacity. 
The overhead incurred by slow start depends on the RTT and bandwidth-delay product of the path. When downloading chunks with large sizes, the probing overhead is minimal but it can be significant for small chunks. Therefore, as we observed in the Puffer data, the throughput achieved for a given chunk in this synthetic simulation depends on the size of the chunk. 

\smallskip
\noindent\textbf{Performance Metric:}
We compare \CausalSim predictions with ground truth counterfactual trajectories, via the \gls{mse} distance between the two time series:
\begin{equation}
    MSE(\mathbf{p}, \mathbf{q}) = ||\mathbf{p} - \mathbf{q}||_2^2
\end{equation}
Here, $\mathbf{p}=\{p_t\}_{t=1}^N$ and $\mathbf{q}=\{q_t\}_{t=1}^N$ are time series vectors. 
Better predictions yield smaller \gls{mse} values, where an ideal \gls{mse} is 0.

\subsubsection{Data \& Algorithms}
\label{app:abr_synth:alg_data}

\begin{table*}[!bt]
\small
\centering
\begin{tabular}{ c l l c c }
\toprule
\textbf{Policies} & \textbf{Hyperparameter} & \textbf{Value} & Used as source & Used as left out \\
\midrule
\multirow{2}{*}{\textbf{\gls{bba}}} & Cushion & 5 & \multirow{2}{*}{\checkmark} &  \multirow{2}{*}{\checkmark} \\
\cmidrule{2-3}
& Reservoir & 10 & &\\
\midrule
\multirow{3}{*}{\textbf{BOLA-BASIC}} & $V$ & 0.71 (Computed using puffer formula) & \multirow{3}{*}{\checkmark} &  \multirow{3}{*}{\checkmark} \\
\cmidrule{2-3}
& $\gamma$ & 0.22 (Computed using puffer formula) & &\\
\cmidrule{2-3}
& Utility function & $\ln(\text{chunk sizes})$ (As used in BOLA paper\cite{bola}) & &\\
\midrule
\textbf{Random} & - & - & \checkmark &  \checkmark \\
\midrule
\multirow{3}{*}{\textbf{\gls{bba}-Random mixture 1}} & Cushion & 5 & \multirow{3}{*}{\checkmark} & \multirow{3}{*}{\checkmark} \\
\cmidrule{2-3}
& Reservoir & 10 & &\\
\cmidrule{2-3}
& Random choices & 50\% & &\\
\midrule
\multirow{3}{*}{\textbf{\gls{bba}-Random mixture 2}} & Cushion & 10 & \multirow{3}{*}{\checkmark} & \multirow{3}{*}{\checkmark} \\
\cmidrule{2-3}
& Reservoir & 20 & &\\
\cmidrule{2-3}
& Random choices & 50\% & &\\
\midrule
\multirow{4}{*}{\textbf{MPC}} & Lookback length & 5 & \multirow{4}{*}{\checkmark} & \multirow{4}{*}{\checkmark} \\
\cmidrule{2-3}
& Lookahead length & 5 & &\\
\cmidrule{2-3}
& Rebuffer penalty & 4.3 & &\\
\cmidrule{2-3}
& Throughput estimate & Harmonic mean & &\\
\midrule
\multirow{2}{*}{\textbf{Rate-based}} & Lookback length & 5 & \multirow{2}{*}{\checkmark} & \multirow{2}{*}{\checkmark} \\
\cmidrule{2-3}
& Throughput estimate & Harmonic mean & &\\
\midrule
\multirow{2}{*}{\textbf{Optimistic Rate-based}} & Lookback length & 5 & \multirow{2}{*}{\checkmark} & \multirow{2}{*}{\checkmark} \\
\cmidrule{2-3}
& Throughput estimate & Max & &\\
\midrule
\multirow{2}{*}{\textbf{Pessimistic Rate-based}} & Lookback length & 5 & \multirow{2}{*}{\checkmark} & \multirow{2}{*}{\checkmark} \\
\cmidrule{2-3}
& Throughput estimate & Min & &\\
\midrule
\bottomrule
\end{tabular}
\caption{\gls{abr} algorithms used in the synthetic \gls{abr} experiments.}
\label{table:abr_synth_pol}
\end{table*}

Simulating a trajectory in our synthetic \gls{abr} environment needs three components:
\begin{itemize}
    \item A video, with several bit-rates available. We use "Envivio-Dash3" from the DASH-246 JavaScript reference client~\cite{dash}.
    \item An \gls{abr} algorithm. We have a set of 9 policies to choose from, presented in \Cref{table:abr_synth_pol}.
    \item A network path, which is characterized by the latent network capacity and the path \gls{rtt}. 
\end{itemize}

We use random generative processes to generate 5000 network traces and \glspl{rtt}. The \gls{rtt} for a streaming session is sampled randomly, according to a uniform distribution:
\begin{equation*}
    rtt~\sim~Unif(10~ms,~500~ms)
\end{equation*}
Our trace generator is a bounded Gaussian distribution, whose mean comes from a Markov chain. Prior work shows Markov chains are appropriate models for TCP throughput~\cite{cs2p}, and Gaussian distributions can model throughputs in stationary segments of TCP flows~\cite{tcp_gauss}.

Concretely, at the start of the trace, the following parameters are randomly sampled:
\begin{align*}
    v~&\sim~Unif(30,~100) \\
    p~&=~1/v \\
    l,~h~&\sim~ Unif(0.5,~4.5)\\
    &s.t.~\frac{h-l}{h+l}~>~0.3\\
    s_0~&\sim~Unif(l,~h) \\
    c_{\sigma}~&\sim~Unif(0.05,~0.3)
\end{align*}
At each time step, the state remains unchanged with probability $1-p$ and changes otherwise. When changing, the next state is sampled from a double exponential distribution centered around the previous state:
\begin{align*}
    \lambda~&=~solve_{x \in \mathbb{R}^+}(1-~e^{x(h-s_{t-1})}-~e^{x(s_{t-1}-l)} = 0) \\
    s_t~&=~DoubleExp(s_{t-1},~\lambda)
\end{align*}
The point for this specific transition kernel is that small changes in network capacity should be more likely than drastic changes.
Finally, the network capacity $c_t$ in each step is sampled from a Gaussian distribution, defined by these parameters:
\begin{equation*}
    c_t~\sim~Normal(s_t,~s_t \cdot c_{\sigma})
\end{equation*}

\subsubsection{Training setup}

\begin{table*}
\small
\centering
\begin{tabular}{ c l l }
\toprule
\textbf{Model} & \textbf{Hyperparameter} & \textbf{Value} \\
\midrule
& Hidden layers (SLSim) & (128, 128) \\
\cmidrule{2-3}
 & Hidden layers (\CausalSim: Extractor, Discriminator and $\mathcal{F}_{system}$) & (128, 128) \\
\cmidrule{2-3}
& Hidden layers (\CausalSim: Action encoder) & (64, 64) \\
\cmidrule{2-3}
& Rank $r$ & 2 \\
\cmidrule{2-3}
\textbf{\CausalSim} (4 networks)  & Hidden layer Activation function & \gls{relu} \\
\cmidrule{2-3}
& Output layer Activation function & Identity mapping \\
\cmidrule{2-3}
& Optimizer & Adam \cite{adam} \\
\cmidrule{2-3}
\textbf{SLSim} (1 network)  & Learning rate & 0.0001 \\
\cmidrule{2-3}
& $\beta_1$ & 0.9 \\
\cmidrule{2-3}
& $\beta_2$ & 0.999 \\
\cmidrule{2-3}
& $\epsilon$ & $10^{-8}$ \\
\cmidrule{2-3}
& Batch size & $2^{13}$ \\
\midrule
\multirow{5}{*}{\textbf{\CausalSim}} & $\kappa$ & \{0.01, 0.1, 1, 10, 100\} \\
\cmidrule{2-3}
& Training iterations (num\_train\_it) & 20000 \\
\cmidrule{2-3}
& num\_disc\_it & 10 \\
\cmidrule{2-3}
& Loss function & \{MSE\} \\
\midrule
\textbf{SLSim} & Training iterations & 20000 \\
\cmidrule{2-3}
& Loss function & \{Huber($\delta=1.0$), L1, MSE\} \\
\midrule
\bottomrule
\end{tabular}
\caption{Training setup and hyperparameters for the synthetic \gls{abr} experiments.}
\label{table:abr_synth_train}
\end{table*}

Similar to the real-world \gls{abr} experiment, we use \glspl{mlp} as the \gls{nn} structures for \CausalSim models and the SLSim model. We tune all the hyperparameters of both baselines as is done in the real-world \gls{abr} experiment (see \S\ref{sec:puffer_hyper_params} and \S\ref{sec:slsim_hyper_params}). \Cref{table:abr_synth_train} comprehensively lists all hyperparameters used in training.

\subsection{Can \CausalSim Faithfully Simulate  \\ New Policies?}
\label{app:abr_synth:extended_eval}
\begin{figure*}[!bt]
\vspace{0.5cm}
  \begin{subfigure}{0.28\textwidth}
    \centering
    \hspace*{-0.03\linewidth}
    \begin{tikzpicture}
        \begin{axis}[height=5cm,
                     width=\linewidth,
                     axis lines=left,
                     xlabel={MSE},
                     ylabel={CDF (\%)},
                     font=\small,
                     legend columns=1,
                     enlargelimits=true,
                     legend style={at={(0.4,0.2)}, 
                          anchor=west, fill=white, fill opacity=0.7, draw opacity=1, 
                          text opacity=1, draw=none, font=\tiny, inner sep=0},
                          legend cell align={left},
                     xmax=30,
                     xmin=0,
                     ymax=100, 
                     ymin=0,
                     ytick={10, 30, 50, 70, 90},
                     xtick={0, 10, 20, 30},
                     ylabel near ticks,
                     xlabel near ticks,
                     grid=both
                     ]
            \input{figures/raw_data/abr_synth/buffer_mse_cdf}
            \legend{\CausalSim predictions, ExpertSim predictions, SLSim predictions};
        \end{axis}
    \end{tikzpicture}
    \caption{}
    \label{fig:abr_synth:emd_cdf}
  \end{subfigure}
  \begin{subfigure}{0.28\textwidth}
    \centering
    \hspace*{-0.03\linewidth}
    \begin{tikzpicture}
        \begin{axis}[height=5cm,
                     width=\linewidth,
                     axis lines=left,
                     xlabel={MSE},
                     ylabel={CDF (\%)},
                     font=\small,
                     legend columns=1,
                     enlargelimits=true,
                     legend style={at={(0.4,0.2)}, 
                          anchor=west, fill=white, fill opacity=0.7, draw opacity=1, 
                          text opacity=1, draw=none, font=\tiny, inner sep=0},
                          legend cell align={left},
                     xmax=2,
                     xmin=0,
                     ymax=100, 
                     ymin=0,
                     ytick={10, 30, 50, 70, 90},
                     xtick={0, 0.5, 1, 1.5, 2},
                     ylabel near ticks,
                     xlabel near ticks,
                     grid=both
                     ]
            \input{figures/raw_data/abr_synth/buffer_mse_cdf}
            \legend{\CausalSim predictions, ExpertSim predictions, SLSim predictions};
        \end{axis}
    \end{tikzpicture}
    \caption{}
    \label{fig:abr_synth:emd_cdf_close}
  \end{subfigure}
    \begin{subfigure}{0.43\textwidth}
    \centering
        \begin{tikzpicture}
            \begin{axis}[
            colorbar,
            colorbar  style={
                ylabel={Population (\%)}},
            colormap name=whiteblack,
            point meta max=3,
            point meta min=0,
            tick align=outside,
            tick pos=both,
            x grid style={white!69.0196078431373!black},
            xlabel={Ground truth},
            ymin=3.75, ymax=10.25,
            xmin=3.75, xmax=10.25,
            xtick style={color=black},
            xtick style={color=black},
            y grid style={white!69.0196078431373!black},
            ytick style={color=black},
            height=5cm,
            width=0.7\linewidth,
            font=\small,
            axis lines=left,
            xlabel={Ground Truth},
            ylabel={\CausalSim's Predictions},
            ytick style={color=black},
            ylabel near ticks,
            xlabel near ticks,
            ]
            \addplot graphics [includegraphics cmd=\pgfimage,xmin=3.5, xmax=10.25, ymin=3.5, ymax=10.25] {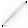};
            \end{axis}
        \end{tikzpicture}
    \caption{}
    \label{fig:abr_synth:heatmap}
  \end{subfigure}
  \caption{\textbf{(a)} Distribution of \CausalSim, ExpertSim, and SLSim \gls{mse}s over all possible source left-out pairs. \textbf{(b)} The same figure with a smaller MSE range. In this magnified view, \CausalSim clearly outperforms the baselines. \textbf{(c)} Two-dimensional histogram heatmap of \CausalSim predictions vs. ground truth.}
\end{figure*}

Similar to our real-data evaluations, we train models based on training data generated using all policies except a left-out policy, for which the model does not observe any data. 
Although traces come from the same generative process, no two trajectories in the dataset collected with different policies share the exact same trace, as this would be an unrealistic data collection scenario. 
Given that we have 9 possible policies to leave out, we have 9 possible datasets and models.
There are 8 possible groups of trajectories to choose as sources, based on the policy that generated them. 
In total this leaves 72 different combinations and scenarios. 
We use the same hyper-parameter tuning approach examined in \S\ref{sec:puffer_hyper_params}. \Cref{fig:abr_synth:emd_cdf} compares the CDF of \gls{mse} values resulting from \CausalSim and the two baselines. 
As evident, both baselines suffer from inaccurate predictions and in some cases are catastrophically inaccurate. On the contrary, \CausalSim maintains favorable performance, even in the tail of its \gls{mse} distribution. 
\Cref{fig:abr_synth:emd_cdf_close} gives a closer look at the CDF curves. 
We see \CausalSim dominates at every scale.

\Cref{fig:abr_synth:heatmap} is a heatmap of the two dimensional histogram of \CausalSim predictions and ground truths. 
A fully accurate prediction scheme would perfectly match the ground truth and only the diagonal of this histogram would be populated.
\CausalSim almost achieves that, indicating it produces accurate trajectories on a step-by-step basis. 

\begin{figure}[!ht]
    \begin{tikzpicture}
        \begin{axis}[height=5cm,
                     width=\linewidth,
                     axis lines=left,
                     xlabel={Chunk index},
                     ylabel={MAPE (\%)},
                     font=\small,
                     legend columns=1,
                     enlargelimits=true,
                     legend style={at={(0.4,0.2)}, 
                          anchor=west, fill=white, fill opacity=0.7, draw opacity=1, 
                          text opacity=1, draw=none, font=\tiny, inner sep=0},
                          legend cell align={left},
                     xmax=30,
                     xmin=0,
                     ymax=12, 
                     ymin=0,
                     ylabel near ticks,
                     xlabel near ticks,
                     grid=both
                     ]
            \input{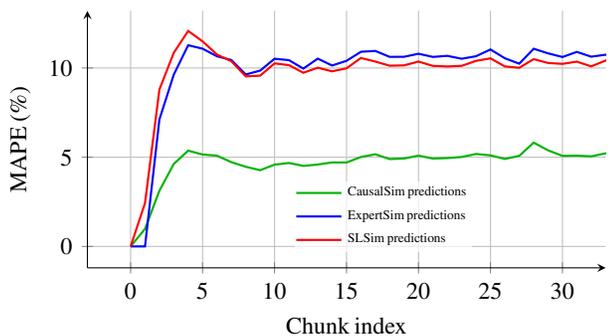}
            \legend{\CausalSim predictions, ExpertSim predictions, SLSim predictions};
        \end{axis}
    \end{tikzpicture}
    \caption{A time series plot of the Mean Absolute Percentage Error (MAPE) across all trajectories, for \CausalSim, ExpertSim and SLSim predictions. Notice how errors accumulate in trajectory simulation.}
    \label{fig:abr_synth:timeseries}
\end{figure}

Further, in \Cref{fig:abr_synth:timeseries}, we compare the the Mean Absolute Percentage Error (MAPE) of \CausalSim, ExpertSim and SLSim predictions across all trajectories at each time step for the first 35 steps. %
Note that the error naturally accumulates for all three methods as we move froward in time. However, \CausalSim maintains a MAPE of (${\sim}5.1\%$) which significantly lower than both ExpertSim's and SLSim's (${\sim}10\%$).

\subsection{Learning ABR policies with \CausalSim}
\label{app:subsec:abr_rl}

\begin{figure*}[!t]
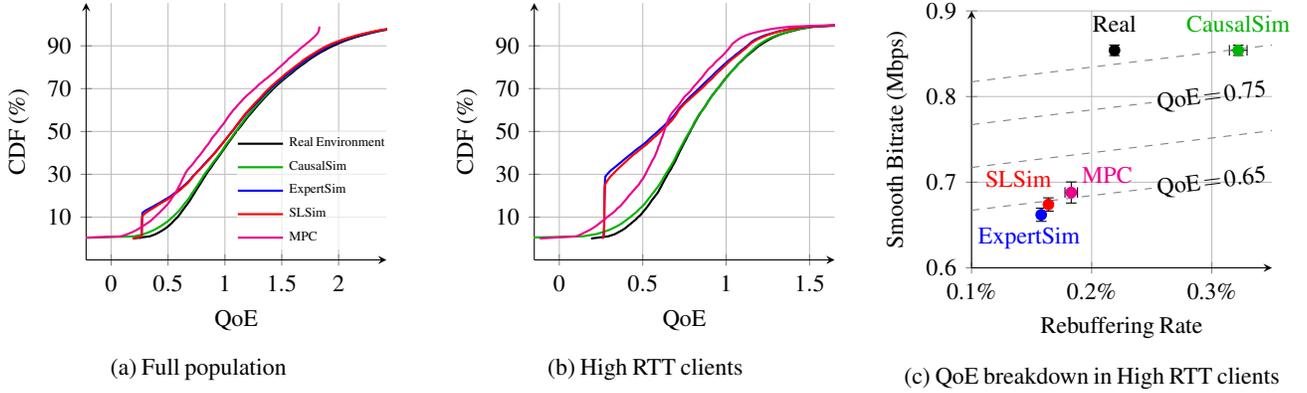

  \begin{subfigure}{0.33\textwidth}
    \centering
    \hspace*{-0.03\linewidth}
    \begin{tikzpicture}
        \begin{axis}[height=5cm,
                     width=0.95\linewidth,
                     axis lines=left,
                     xlabel={QoE},
                     ylabel={CDF (\%)},
                     font=\small,
                     legend columns=1,
                     enlargelimits=true,
                     legend style={at={(0.5, 0.5)}, 
                          anchor=north west, fill=white, fill opacity=0.7, draw opacity=1, 
                          text opacity=1, draw=none, font=\tiny, inner sep=0},
                          legend cell align={left},
                     xmax=2.2,
                     xmin=0,
                     ymax=100, 
                     ymin=0,
                     ytick={10, 30, 50, 70, 90},
                     xtick={0, 0.5, 1, 1.5, 2},
                     ylabel near ticks,
                     xlabel near ticks,
                     grid=both
                     ]
            \input{figures/raw_data/abr_rl/cdf_full_v1}
            \legend{Real Environment, \CausalSim, ExpertSim, SLSim, MPC};
        \end{axis}
    \end{tikzpicture}
    \caption{Full population}
    \label{fig:abr_rl:full_qoe_cdf}
  \end{subfigure}
  \begin{subfigure}{0.33\textwidth}
    \centering
    \hspace*{-0.03\linewidth}
    \begin{tikzpicture}
        \begin{axis}[height=5cm,
                     width=0.95\linewidth,
                     axis lines=left,
                     xlabel={QoE},
                     ylabel={CDF (\%)},
                     font=\small,
                     legend columns=1,
                     enlargelimits=true,
                     legend style={at={(0.6, 0.2)}, 
                          anchor=west, fill=white, fill opacity=0.7, draw opacity=1, 
                          text opacity=1, draw=none, font=\tiny, inner sep=0},
                          legend cell align={left},
                     xmax=1.5,
                     xmin=0,
                     ymax=100, 
                     ymin=0,
                     ytick={10, 30, 50, 70, 90},
                     xtick={0, 0.5, 1, 1.5},
                     ylabel near ticks,
                     xlabel near ticks,
                     grid=both
                     ]
            \input{figures/raw_data/abr_rl/cdf_high_v1}
        \end{axis}
    \end{tikzpicture}
    \caption{High RTT clients}
    \label{fig:abr_rl:high_qoe_cdf}
  \end{subfigure}
  \begin{subfigure}{0.33\textwidth}
    \centering
    \hspace*{-0.03\linewidth}
    \begin{tikzpicture}
        \begin{axis}[
            height=5cm,
            width=0.95\linewidth,
            axis lines=left,
            ylabel near ticks,
            xlabel near ticks,
            xlabel=Rebuffering Rate,
            ylabel=Smooth Bitrate (Mbps),
            grid=both,
            font=\small,
            xmin=0.1,xmax=0.35,
            ymin=0.6, ymax=0.9,
            xtick={0.1, 0.2, 0.3},
            xticklabels={0.1\%, 0.2\%, 0.3\%},
            ]
    
            \addplot[
            scatter/classes={d={black}, a={green!70!black}, c={blue}, b={red}, e={magenta}},
            scatter,
            only marks, 
            visualization depends on=\thisrow{visx} \as \myshiftx,
            visualization depends on=\thisrow{visy} \as \myshifty,
            every node near coord/.append style = {shift={(axis direction
            cs:\myshiftx,\myshifty)}},
            scatter src=explicit symbolic,
            nodes near coords*={\small\Label},
            visualization depends on={value \thisrow{label} \as \Label},
            ]
            plot [error bars/.cd, y dir = both, y explicit, x dir = both, x explicit]
            table[meta=class, x=x, y=y, y error=ey, x error=ex]{
                x   y   ex  ey  visx    visy  class   label
                0.219	0.854	0.00267	0.00617 0   0.01	d	Real
                0.322	0.854	0.00736	0.006   0   0.01	a	\textcolor{green!70!black}{\CausalSim}
                0.158	0.662	0.000867	0.00771 -0.01   -0.05	c	\textcolor{blue}{ExpertSim}
                0.164	0.674	0.00132	0.00775   -0.025   0.01	b	\textcolor{red}{SLSim}
                0.183	0.688	0.00521	0.0123  0.03    0	e	\textcolor{magenta}{MPC}
            };
            
            \addplot[dashed, gray, mark=none] coordinates{
                (0.1, 0.6672)
                (0.25, 0.6930)
            };
            
            \addplot[dashed, gray, mark=none] coordinates{
                (0.1, 0.7172)
                (0.4, 0.7688)
            };
            
            \addplot[dashed, gray, mark=none] coordinates{
                (0.1, 0.7672)
                (0.25, 0.7930)
            };
            
            \addplot[dashed, gray, mark=none] coordinates{
                (0.1, 0.8172)
                (0.4, 0.8688)
            };
            
            \node[rotate=7] (text1) at (axis cs:0.3,0.70){$\text{QoE}=0.65$};
            
            \node[rotate=7] (text2) at (axis cs:0.3,0.80){$\text{QoE}=0.75$};
            
        \end{axis}
    \end{tikzpicture}
    \caption{QoE breakdown in High RTT clients}
    \label{fig:abr_rl:high_qoe_break}
  \end{subfigure}
  \caption{\small \CausalSim trained policies perform well, only marginally behind training on the real environment. Distribution of \gls{qoe} in policies trained with the real environment, \CausalSim, ExpertSim, and the MPC policy. \CausalSim does not underestimate bandwidth in high RTT clients and trains policies that strike the best balance in QoE goals.}
\end{figure*}

We observed how \CausalSim can be used to design an improved policy in \S\ref{subsec:puffer_bayes}, and verified this through deployment in the wild. We would like to take these experiments one step further and ask \emph{can \CausalSim be used to design learning-based policies, such as with \gls{rl}?}

Recent work has shown that \gls{rl} algorithms can learn strong ABR policies by learning through interactions with the environment~\cite{pensieve}. Could we use a \CausalSim model to train high-performance ABR policies without direct environment interaction?
As a first step, we decided to carry out an initial experiment in the synthetic \gls{abr} environment.
We build a \CausalSim model using traces from a ``simulated RCT'' on the synthetic environment.

\smallskip
\noindent\textbf{Performance Metric.}
ABR algorithms are typically evaluated through \gls{qoe} metrics~\cite{mpc}. Assuming the chosen bitrate at step $t$ was $q_t$, the download time was $d_t$ and the buffer was $b_t$, we use the following \gls{qoe} definition:
$$
QoE_t = q_t - |q_{t}-q_{t-1}| - \mu \cdot max(0, d_t-b_{t-1})
$$

This \gls{qoe} metric captures three goals (in succession): 1) Stream in high quality, 2) Maintain a stable quality, 3) Avoid rebuffering. Better policies yield higher \gls{qoe} values, where an ideal \gls{qoe} is equal to the max bitrate.

\subsubsection{How to train policies via simulators?}
\label{app:subsub:train_with_simulators}
To train the \gls{rl} agent, we take a set of logged trajectories where the source policy was MPC and feed them to \CausalSim. In each step, \CausalSim will predict the next counterfactual observation and reward, and the \gls{rl} agent will choose the next counterfactual action based on that observation. This process repeats until this simulated session is over, after which the counterfactual trajectory is used to train the \gls{rl} agent. 
For the \gls{rl} algorithm, we utilize the \gls{a2c} method, a prominent on-policy algorithm, along with \gls{gae}.
\Cref{table:abr_rl_train} lists all hyperparameters for the \gls{rl} training.

\begin{table*}
\small
\centering
\begin{tabular}{c l l }
\toprule
\textbf{Group} & \textbf{Hyperparameter} & \textbf{Value} \\
\midrule

\multirow{14}{*}{Neural Network} & Hidden layers & (32, 32) \\
\cmidrule{2-3}
& Hidden layer activation function & ReLU \\
\cmidrule{2-3}
& \multirow{2}{*}{Output layer activation function} & \gls{a2c} actor: Softmax \\
\cmidrule{3-3}
& & \gls{a2c} critic: Identity mapping\\
\cmidrule{2-3}
& Optimizer & Adam \cite{adam} \\
\cmidrule{2-3}
& Learning rate & 0.001 \\
\cmidrule{2-3}
& $\beta_1$ & 0.9 \\
\cmidrule{2-3}
& $\beta_2$ & 0.999 \\
\cmidrule{2-3}
& $\epsilon$ & $10^{-8}$ \\
\cmidrule{2-3}
& Weight decay & $10^{-4}$ \\
\midrule

\multirow{8}{*}{\gls{a2c} training} & Episode lengths & 490 \\
\cmidrule{2-3}
& Epochs to convergence ($T_c)$ & 8000 (3920000 samples) \\
\cmidrule{2-3}
& Random seeds & 4 \\
\cmidrule{2-3}
& $\gamma$ & 0.96 \\
\cmidrule{2-3}
 & Entropy schedule & 0.1 to 0 in 5000 epochs \\
\cmidrule{2-3}
& $\lambda$ (for \gls{gae}) & 0.95 \\
\midrule

\multirow{2}{*}{Environment} & Chunk length $c$ & 4 \\
\cmidrule{2-3}
& Number of actions (bitrates) & 6 \\
\midrule
\bottomrule
\end{tabular}
\caption{Training setup and hyperparameters for learning RL policies in the synthetic \gls{abr} environment.}
\label{table:abr_rl_train}
\end{table*}

\subsubsection{Does \CausalSim train better policies?}
\label{app:subsub:rl_abr_results}
\Cref{fig:abr_rl:full_qoe_cdf} plots the CDF of average session \gls{qoe} that each policy attains. Here, \emph{Real Environment} refers to training directly with the synthetic \gls{abr} environment, and \CausalSim, ExpertSim and SLSim refer to policies trained by using each of these simulators. \CausalSim trains policies nearly as well as training directly on the environment, while ExpertSim and SLSim fail to provide robust policies across all sessions. \Cref{fig:abr_rl:high_qoe_cdf} plots the CDFs for the high RTT (above 300 ms) clients, where the gap between \CausalSim and the baseline simulators is even larger. 

In this environment, chunk are downloaded according to the slow start model, where congestion control must ramp up its window size over several RTTs before the download rate can reach the available bandwidth. As a result, downloads of smaller chunks (with lower bitrates) incur a noticeable overhead, particularly on high-RTT paths. This overhead becomes less apparent as chosen bitrates become larger. Biased simulators such as SLSim and ExpertSim, which assume all actions lead to the same observed bandwidth, overestimate the achieved rate when counterfactual bitrates are smaller than factual ones (chosen by the source policy) and underestimate it when the counterfactual bitrates are larger. Since the source policy is conservative and tends to choose low bitrates, ExpertSim and SLSim find larger bitrates to be undesirable in the \gls{qoe} trade-off. This can be seen in \Cref{fig:abr_rl:high_qoe_break}, which visualizes the 3 aspects of \gls{qoe} in terms of the rebuffering rate and the smoothed birate, i.e the chosen bitrates with the smoothnes penalty. Notice how policies trained on the real environment and \CausalSim utilize the network by $200$~ kbps more than other policies. The extra rebuffering that \CausalSim incurs is negligible compared to the extra bitrate: 5.9 seconds every hour.

\subsection{Low-rank structure}
\label{app:synth:low_rank}

As discussed in \S\ref{sub:matcomp_cf}, we can formulate the counterfactual estimation problem in the context of matrix completion. For each time step, we know the chosen bitrate (action) and the achieved throughput (trace). We also know the trace is computed using a latent factor and the action.
Suppose the latent factor is the network bottleneck capacity $c_t$\footnote{There may be other latent factors but bottleneck capacity is likely to have the strongest influence on the achieved throughput.}. $\mathcal{F}_{\text{trace}}$ describes how the achieved throughput (the trace) relates to this latent factor. Intuitively, this should be a close-to-linear function, $m_t \approx c_t$. But it's not exactly linear; for example, congestion control may under-utilize the network capacity for small transfers on high-RTT paths. 

We form a matrix $M$, where the rows denote actions $a_t \in [A]$ and the columns denote the latent factors $u_t^i$ for each trajectory. The `factual' data we have are single observed trace values in each column, i.e for each step and each latent, we have observed the trace from a single action. To estimate counterfactuals, we must complete the matrix.  
We have no way of knowing the true $\mathcal{F}_{\text{trace}}$ in the Puffer dataset. But to get a sense for what it might look like and whether it's plausible that $M$ is low rank, we can investigate this in the synthetic \gls{abr} environment instead. 

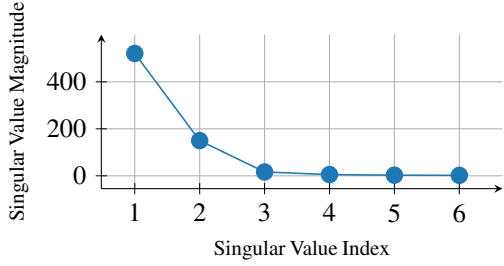
\begin{figure}
    \centering
    \hspace*{-1cm}
    \begin{tikzpicture}
    \definecolor{color0}{rgb}{0.12156862745098,0.466666666666667,0.705882352941177}
    \begin{axis}[
    height = 5cm,
    width = \linewidth + 2cm,
    tick align=outside,
    tick pos=both,
    grid = both,
    axis lines = left,
    x grid style={white!69.0196078431373!black},
    xlabel={\footnotesize Singular Value Index},
    xmin=1, xmax=6.15,
    xtick style={color=black},
    xtick = {1, 2, 3, 4, 5, 6},
    y grid style={white!69.0196078431373!black},
    ylabel={\footnotesize Singular Value Magnitude},
    ymin=0, ymax=546.717969647616,
    scale=0.6,
    enlargelimits=true,
    ytick style={color=black},
    ytick = {0, 200, 400}
    ]
    \addplot [semithick, color0, mark=*, mark size=3, mark options={solid}]
    table {
    1 520.777478068525
    2 149.954706544678
    3 16.3032793251514
    4 5.2349692334538
    5 2.70755782219104
    6 1.9676464866997
    };
    \end{axis}
    \end{tikzpicture}
    \caption{Singular values of matrix $M$ in synthetic \gls{abr} suggest that $M$ is approximately rank $2$.}
    \label{fig:low_rank_ABR}
\end{figure}

For the TCP slow start model this environment uses, $\mathcal{F}_{\text{trace}}$ takes the following form:

\begin{align}
&\text{Let} ~~~ \hat{RTT} \coloneqq \frac{RTT}{\ln(2)}\\
    m_t =& 
    \begin{dcases}
        \frac{c_t}{1 + \frac{\hat{RTT} \cdot (\ln(c_t/\dot{c})-c_t+\dot{c})}{s_t}} & \text{if } s_t \ge \hat{RTT}.(c_t-\dot{c})\\
        \frac{s_t}{\hat{RTT} \cdot \ln(\frac{s_t}{\hat{RTT} \cdot \dot{c}}+1)}              & \text{otherwise}
    \end{dcases}
\end{align}

where $s_t$ is the chunk size (which itself is determined by the bitrate chosen by ABR) and $\dot{c}$ is the starting download rate in the slow start algorithm (in our case, equal to 2 MTUs).
We use this model to generate a version of $M$ with 
$A=6$ actions and 
$U = 49000$ latent network conditions. 
We compute the singular value decomposition with the 6 singular values represented in non-increasing order ($\sigma_1 \geq \sigma_2 \geq \cdots \geq \sigma_6$). 
The total ``energy'' of matrix is given by sum of squares of these singular values. It turns out that $\frac{\sigma_1^2 + \sigma_2^2}{\text{total energy}}$ 
is more than $0.999$. This suggests that most of the matrix is captured by its rank-2 approximation, as depicted in \Cref{fig:low_rank_ABR}. 
In other words, $M$ is approximately low ($=2$) rank. 


\section{Load Balancing}
\label{app:lb}

\subsection{Does \CausalSim Faithfully Infer Latent States?}\label{app:lb_lat_match}
\begin{figure}
    \centering
        \begin{tikzpicture}
        \begin{axis}[
            colorbar, 
            colormap name=whiteblack,
            point meta min=0.0,
            point meta max=5000,
            colorbar style={
                ylabel={Population count},
                ytick={0, 1000, 2000, 3000, 4000, 5000}
            },
            height=4cm,
            width=0.55\linewidth,
            axis lines=left,
            xlabel={Latent job size},
            font=\small,
            enlargelimits=true,
            xmin=0, xmax=1200,
            xmin=0, xmax=1300,
            ymin=0.1, ymax=20,
            xtick style={color=black},
            y grid style={white!69.0196078431373!black},
            ylabel={\CausalSim's extracted feature},
            ylabel near ticks,
            xlabel near ticks,
             ]
            \addplot[thick,blue] graphics[xmin=0, xmax=1300, ymin=0.1, ymax=20,] {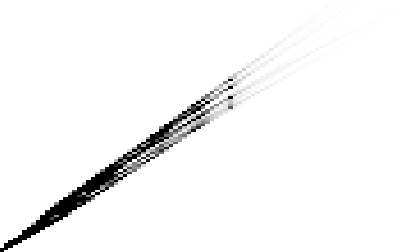};
        \end{axis}
    \end{tikzpicture}
    \vspace{-4mm}
    \caption{Two-dimensional histogram heatmap of \CausalSim extracted latent state vs. latent job sizes.}
    \label{fig:lb:features}
    \vspace{-4mm}
  \end{figure}

We test the claim that estimating the exogenous latent state and using it to predict the next state was indeed the key to producing accurate counterfactual predictions, as the architecture of \CausalSim suggests. 
To do so, we compare \CausalSim's estimated latent state with the underlying job sizes---the job size is indeed the latent state that dictates the dynamics in the load balancing environment.
We find that the estimated latent states and the job sizes are highly correlated, as illustrated in \Cref{fig:lb:features}, with a \gls{pcc} of 0.994. 
This demonstrates that \CausalSim can learn faithful representations of true latent states.

\subsection{Data \& Algorithms}
\label{app:lb:alg_data}

\begin{table*}[!bt]
    \small
    \centering
    \begin{tabular}{ c l c c }
    \toprule
    \textbf{Policies} & \textbf{Description} & Used as source & Used as left out \\
    \midrule
    Server limited policy (8 variations) & Randomly assign to only two servers & \checkmark & $\times$ \\
    \midrule
    Shortest queue & Assign to server with smallest queue & \checkmark & \checkmark\\
    \midrule
    Power of $k$ ($k \in \{2, 3, 4, 5\})$ & Poll queue lengths of $k$ server and assign to shortest queue & \checkmark & \checkmark\\
    \midrule
    \multirow{2}{*}{Oracle optimal} & Normalize queue sizes with server rates & \multirow{2}{*}{\checkmark} & \multirow{2}{*}{\checkmark} \\
     & and assign to shortest normalized queue & & \\
    \midrule
    \multirow{2}{*}{Tracker optimal} & Similar to oracle, but estimates server rates & \multirow{2}{*}{\checkmark} & \multirow{2}{*}{\checkmark} \\
     & with historical observations of processing times & & \\
    \midrule
    \bottomrule
    \end{tabular}
    \caption{Scheduling policies used in the load balancing experiment.}
    \label{table:lb_pol}
\end{table*}

To simulate the load balancing problem described in \S\ref{lb:exp_setup}, we need to set the server processing rates $\{r_i\}_{i=1}^N$, and arriving job sizes $S_k$. Server rates are generated randomly, as follows:
\begin{align}
    r_i~&=~e^{u_i} \\
    \text{where~}~u_i~&\sim~Unif(-\ln(5), \ln(5)) 
\end{align}

We generate job sizes using a time-varying Gaussian distribution. At step $k$ of the trajectory, job size $S_k$ is sampled as follows:
\begin{equation*}
    S_k \sim Normal(\mu_k, \sigma_k)
\end{equation*}
where $\mu_k$ and $\sigma_k$ signify the mean and variance of the generative distribution at time step $k$. At each time step, with a probability of $p=1/12000$, the mean and variance change and with a probability of $1-p$, they remain the same. The mean and variance values are drawn from random distributions, both at the start of a trajectory and when a change occurs, in the following manner:

\begin{align}
    \text{If~} k=&0 \text{ (start of trace) or, mean and variance must change:} \nonumber\\
    \mu_k &\sim Pareto(\alpha=1,~~L=10^1,~~H=10^{2.5}) \\
    \sigma_k &\sim Unif(0,~~0.5 \mu_k) \\
    \text{Else:}& \nonumber\\
    \mu_k&= \mu_{k-1} \\
    \sigma_k &= \sigma_{k-1}
\end{align}

Jobs generated according to this process are temporally correlated, and therefore not independent and identically distributed. Training data consists of 5000 trajectories of length 1000, each of which was randomly assigned a policy from a set of 16 policies, described in \Cref{table:lb_pol}.

\subsection{Training setup}

\begin{table*}
\small
\centering
\begin{tabular}{ c l l }
\toprule
\textbf{Model} & \textbf{Hyperparameter} & \textbf{Value} \\
\midrule
& Hidden layers (SLSim) & (128, 128) \\
\cmidrule{2-3}
 & Hidden layers (\CausalSim: Extractor, Discriminator) & (128, 128) \\
\cmidrule{2-3}
& Hidden layers (\CausalSim: Action encoder) & No hidden layers \\
\cmidrule{2-3}
& Rank $r$ & 1 \\
\cmidrule{2-3}
\textbf{\CausalSim} (3 networks) & Hidden layer Activation function & \gls{relu} \\
\cmidrule{2-3}
& Output layer Activation function & Identity mapping \\
\cmidrule{2-3}
& Optimizer & Adam \cite{adam} \\
\cmidrule{2-3}
\textbf{SLSim} (1 network) & Learning rate & 0.0001 \\
\cmidrule{2-3}
& $\beta_1$ & 0.9 \\
\cmidrule{2-3}
& $\beta_2$ & 0.999 \\
\cmidrule{2-3}
& $\epsilon$ & $10^{-8}$ \\
\cmidrule{2-3}
& Batch size & $2^{13}$ \\
\midrule
\multirow{3}{*}{\textbf{\CausalSim}} & $\kappa$ & \{0.01, 0.1, 1, 10, 100\} \\
\cmidrule{2-3}
& Training iterations (num\_train\_it) & 10000 \\
\cmidrule{2-3}
& num\_disc\_it & 10 \\
\midrule
\textbf{SLSim} & Training iterations & 10000 \\
\cmidrule{2-3}
& Loss function & Huber, L1, MSE \\
\midrule
\bottomrule
\end{tabular}
\caption{Training setup and hyperparameters for the load balancing experiment.}
\label{table:lb_train}
\end{table*}

As before, we use \glspl{mlp} as the \gls{nn} structures for \CausalSim models and the SLSim model and \Cref{table:lb_train} is a comprehensive list of all hyperparameters used in training. We tune the parameter $\kappa$ for \CausalSim and the loss function in SLSim in a similar fashion to what is described in \S\ref{sec:puffer_hyper_params} and \S\ref{sec:slsim_hyper_params}.
Note that, as mentioned in \S\ref{lb:exp_setup}, we assume access to $\mathcal{F}_{\text{system}}$ and focus on the more challenging task of estimating the trace quantities, for both \CausalSim and SLSim. 
Therefore, in training, there are no observations and hence $\mathcal{L}_{\text{total}}$ consist of two terms: the squared loss of the trace quantities and the discriminator loss.

\section{Causal Inference Related Work}
\label{app:related}

Identifying causal relationships from observational data is a critical problem in many domains~\cite{guo2020survey}, including medicine\cite{cross2013identification}, epidemiology~\cite{robins2000marginal},  economics~\cite{imbens2004nonparametric}, and education \cite{dehejia1999causal}.
Indeed, identifying causal structure and answering causal inference queries is an emerging theme in different machine learning tasks recently, including computer vision \cite{zhang2020causal,yang2019me}, reinforcement learning \cite{forney2017counterfactual,agarwal2021persim}, fairness~\cite{garg2019counterfactual}, and time-series analysis~\cite{agarwal2020multivariate} to name a few. 
One important aspect about causal inference is its ability to answer counterfactual queries.
For such queries, many methods were developed; where some approaches are motivated by  Pearl’s structural causal model \cite{causality-pearl}, and by Rubin’s potential outcome framework \cite{rubin2005causal}.
We refer the interested reader to recent surveys such as \cite{guo2020survey} and references there in for an overview of recent advances in our ability to infer causal relationships from observational data. 

Another related line of work within this literature is synthetic controls and its extension synthetic interventions, which aims to build synthetic trajectories of different units (e.g. individuals, geographic locations) under unseen interventions by appropriately learning across observed trajectories \cite{agarwal2021synthetic, agarwal2019robustness, RSC, mRSC, abadie1, abadie2}. However, these approaches assume a static set of intervention and do not apply to our setting.
\end{appendices}

\end{document}